\documentclass[preprint,5p,times,sort&compress]{elsarticle_v2} 

\usepackage{amsmath}
\usepackage{amsfonts}
\usepackage{amsthm}
\usepackage{amssymb}
\def\bz{\mathbf{z}}
\def\bx{\mathbf{x}}
\def\by{\mathbf{y}}
\def\bj{\mathbf{j}}
\def\bX{\mathbf{X}}
\usepackage{lineno}    
\PassOptionsToPackage{hyphens}{url}
\usepackage{hyperref}  
\hypersetup{colorlinks=true,breaklinks=true}
\usepackage{xcolor} 
\usepackage{booktabs} 
\usepackage{multirow}
\usepackage{siunitx} 
\usepackage{array} 

\journal{Artificial Intelligence in Medicine}

\begin{document}

\begin{frontmatter}

\title{Interpretable machine learning for time-to-event prediction in medicine and healthcare}

\author[uw]{Hubert Baniecki}\ead{h.baniecki@uw.edu.pl}
\author[pw]{Bartlomiej Sobieski}
\author[pw,wum]{Patryk Szatkowski}
\author[pw,wum]{Przemyslaw Bombinski}
\author[uw,pw]{Przemyslaw Biecek}

\affiliation[uw]{organization={University of Warsaw}, city={Warsaw}, country={Poland}}
\affiliation[pw]{organization={Warsaw University of Technology}, city={Warsaw}, country={Poland}}
\affiliation[wum]{organization={Medical University of Warsaw}, city={Warsaw}, country={Poland}}

\begin{abstract}
Time-to-event prediction, e.g. cancer survival analysis or hospital length of stay, is a highly prominent machine learning task in medical and healthcare applications. However, only a few interpretable machine learning methods comply with its challenges. To facilitate a comprehensive explanatory analysis of survival models, we formally introduce time-dependent feature effects and global feature importance explanations. We show how post-hoc interpretation methods allow for finding biases in AI systems predicting length of stay using a novel multi-modal dataset created from 1235 X-ray images with textual radiology reports annotated by human experts. Moreover, we evaluate cancer survival models beyond predictive performance to include the importance of multi-omics feature groups based on a large-scale benchmark comprising 11 datasets from The Cancer Genome Atlas (TCGA). Model developers can use the proposed methods to debug and improve machine learning algorithms, while physicians can discover disease biomarkers and assess their significance. We hope the contributed open data and code resources facilitate future work in the emerging research direction of explainable survival analysis. 
\end{abstract}


\end{frontmatter}

\section{Introduction}\label{sec:introduction}

Advances in the development of machine learning~(ML) algorithms enable successful deployment of artificial intelligence (AI) systems in medicine and healthcare~\citep{rajpurkar2022ai,zhou2023foundation}. Nevertheless, the black-box nature of complex predictive models inhibits their adoption among stakeholders, including physicians and patients. Numerous interpretable machine learning~(IML) algorithms~\citep{stiglic2020interpretability,molnar2020interpretable,biecek2021explanatory}, contained in a broader field of explainable AI~\citep{holzinger2022xai,ooge2022explaining,combi2022manifesto}, have been proposed to overcome this opaqueness debt and effectively increase trust in automated decisions (or find out reasons to distrust thereof).

Only recently, the IML toolbox expanded beyond classification and regression to include post-hoc explanation methods suited for interpreting time-to-event prediction~\citep{kovalev2020survlime,wang2021counterfactual, rad2022extracting,utkin2022survnam,krzyzinski2023survshap}. These are especially important for ML applied in medicine and healthcare, of which survival analysis and survival models are a big part~\citep{terminassian2024explainable,langbein2024interpretable}. 
Notably, including a time dimension in explanations is crucial to accurately interpret the predicted survival function~\citep{krzyzinski2023survshap}. Omitting the time dimension is a major limitation of popular explanation methods like permutation feature importance~\citep{breiman2001random,fisher2019all}, individual conditional expectation~\citep{goldstein2015peeking} and partial dependence plots~\citep{friedman2001greedy}.

\begin{figure*}
    \centering
    \includegraphics[width=0.8\textwidth]{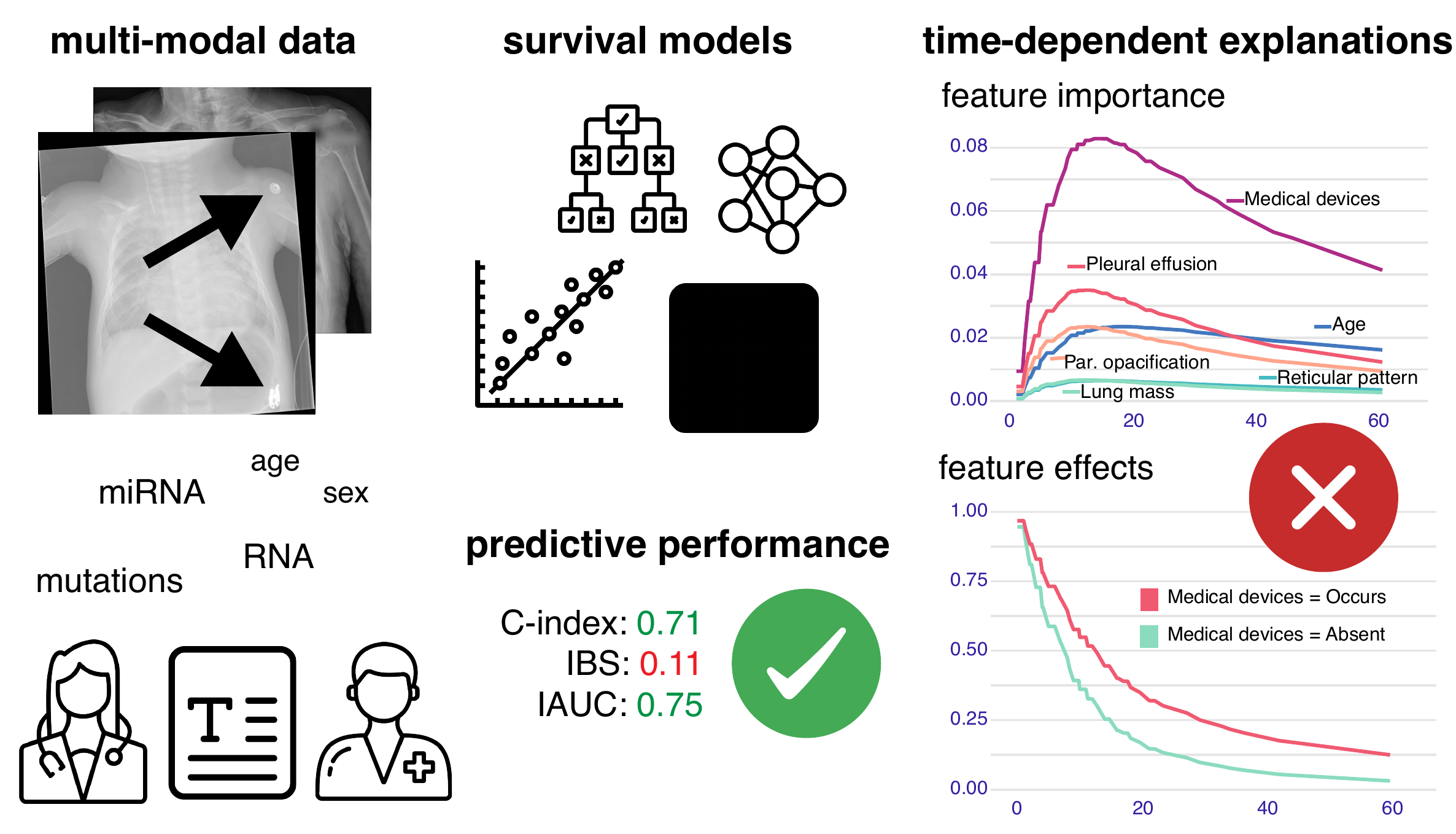}
    \caption{Post-hoc explanation methods allow for finding biases in machine learning models predicting hospital length of stay, and evaluating cancer survival models beyond performance to include the importance of multi-omics feature groups.}
    \label{fig:abstract}
\end{figure*}

To this end, this paper extends our previous work on IML for survival analysis~\citep{baniecki2023hospital} with the following contributions: 
\begin{enumerate}
    \item In Section~\ref{sec:methods}, we formally introduce time-dependent global feature importance and time-dependent feature effects, i.e. model-agnostic explanation methods adapted to the task of time-to-event prediction (Figure~\ref{fig:abstract}).
    \item In Section~\ref{sec:materials-tlos}, we describe in detail the novel dataset for predicting hospital length of stay from X-ray images. We use time-dependent explanations to analyse a ML model predicting hospital length of stay, which exhibits harmful bias towards medical devices appearing in X-ray images.
    \item In Sections~\ref{sec:materials-tcga}~\&~\ref{sec:results-tcga}, we demonstrate the general applicability of the proposed methods through a multi-modal medical use-case illustrating the added value of IML for time-to-event prediction in survival analysis. 
\end{enumerate}
We supplement the paper with an original open-source implementation of time-dependent feature effects and global feature importance in Python.\footnote{\url{https://github.com/mi2datalab/interpret-time-to-event}}

\section{Related work}

\subsection{Interpretable machine learning for survival analysis}

The majority of proposed ML explanation methods omit considering time-to-event analysis~\citep{molnar2020interpretable,biecek2021explanatory}. Yet, ML survival models become more complex with the aim to increase their predictive performance~\citep{polsterl2016survival,jing2019deep,hao2022survivalcnn}. One way of increasing the transparency of predictive outcomes is using the so-called ``interpretable'' model algorithms~\citep{cho2023interpretable,jiang2023decaf,xu2023coxnam}, which resemble simpler interpretations alike statistical models, e.g. Cox proportional hazards. Instead, we relate to research on post-hoc model-agnostic explanations that can be applied to a broad spectrum of black-box learning algorithms. 

SurvLIME~\citep{kovalev2020survlime} is an adaptation of local interpretable model-agnostic explanations~\citep{ribeiro2016should} to survival analysis. In~\citep{wang2021counterfactual}, a counterfactual explanation for predicting the survival of cardiovascular ICU patients is proposed. In~\cite{rad2022extracting}, explanations are developed based on a simple interpretable surrogate model to approximate the decisions of a black-box survival model. In~\citep{utkin2022survnam}, SurvNAM extends the SurvLIME method using general additive models to estimate both local and global explanations. Most recently, SurvSHAP(t)~\cite{krzyzinski2023survshap} is the first method to include the time dimension in explanations of time-to-event prediction. But, it only considers \emph{local} feature attributions and importance.

Currently underdeveloped are time-dependent global methods like permutation feature importance~\citep{breiman2001random,fisher2019all} or partial dependence plots~\citep{friedman2001greedy}, which result from the aggregation of individual conditional expectations~\citep{goldstein2015peeking}. Moreover, none of the explanation methods considers grouping features~\citep{au2022grouped}, which is useful in analysing multi-modal medical data.

While evaluating explanation methods through heuristic performance metrics~\citep{komorowski2023towards} is relevant to theoretical research in AI, our work leans towards the applied side of IML. For example, time-dependent explanations can be used to validate biomarkers of cancer survival~\citep[][figure~3]{donizy2023ki67}. As we have previously shown that juxtaposing different explanation methods increases the accuracy and confidence of human decision making~\citep{baniecki2023grammar}, this paper aims to further facilitate comparative visualization as a useful tool in interpreting ML models in medical and healthcare applications.

\subsection{Predicting and explaining hospital length of stay} 

In Section~\ref{sec:results-tlos}, we demonstrate the applicability of IML methods to explaining hospital length of stay (LoS) predictions as an enabler towards trustworthy human-AI collaboration in radiomics. Predicting patients' hospital LoS is a challenging task supporting the day-to-day decisions of medical doctors and nurses~\cite{huang2013length}. For example, accurate LoS prediction can increase hospital service efficiency, cutting costs and improving patient care. Historically, white-box statistical learning methods were used to estimate the anticipated LoS~\cite{chaou2017predicting}. These provide clear reasoning behind the prediction, which is especially important in medical applications requiring stakeholders to comprehend ``Why?''~\cite{rudin2019stop}. Nowadays, advancements in machine and deep learning for healthcare provide valuable improvements in the performance of predicting LoS~\cite{muhlestein2019predicting,zhang2020combining,wen2022time}. The natural drawback of using not inherently interpretable black-box models is their complex nature~\cite{biecek2021explanatory,rudin2019stop}. Indeed, a recent systematic review on the exact topic of hospital LoS prediction concludes with a concrete statement that there are no studies on the explainability of black-box models predicting LoS~\cite{stone2022systematic}, a matter of high importance for diverse stakeholders involved in this healthcare process. 

Applying AI through ML to predict hospital LoS from data is broadly studied as it has a high potential to support decision making in healthcare~\cite{stone2022systematic}. In~\cite{huang2013length}, LoS is predicted based on information from clinical treatment processes, i.e. sequences of time-point hospital events. In~\cite{chaou2017predicting}, hospital events like the number of tests and time of arrival are aggregated to quantify their importance in patient discharge. We base our analysis on clinical features impacting physician understanding of the patient's severity instead. In~\cite{muhlestein2019predicting}, ML models predict LoS after brain surgery using clinical features in a single-value regression task. This results in simple feature importance and effects explanations without the valuable time dimension. In~\cite{zhang2020combining}, deep learning models classify the patient staying in a hospital for longer than 7 days based on multi-modal data combining unstructured notes and tabular features. In~\citep{wen2022time}, various machine and deep learning survival models are benchmarked for predicting LoS of COVID-19 patients. Both studies focus on predictive performance without explaining the models. For a broader overview of works on LoS prediction, we refer the reader to~\citep{stone2022systematic}.

Contrary to the above-mentioned studies, we specifically use raw X-ray images to analyze the predictive power of radiomics features and explain the prediction in a time-dependent manner. To achieve it, we rely on \texttt{pyradiomics} -- the state-of-the-art radiomics feature extraction tool~\cite{van2017computational} and \texttt{survex} -- a toolbox for explaining ML survival models~\cite{spytek2023survex}.

\section{Methods}\label{sec:methods}

In this section, we formulate an adaptation of IML methods used to explain time-to-event predictions in Section~\ref{sec:results}. Unlike classification or regression tasks, ML survival models output a function in time. 

Let us denote by $f_t(\bx)$ a prediction for an observation $\bx$ at timestep $t$, where $f$ is the function estimated by the model and $t$ ranges from $t_0$ to $T$. Note that $f$ is not restricted to represent the survival function or cumulative hazard function, as the formulation of the following methods is correct in both cases. An observation $\bx$ is a vector of the realizations of $p$ features from a dataset $\bX$ with a total number of $n$ observations. By $\bx^{j|=z}$, we denote an observation in which the value of the $j$-th feature is replaced with $z$ and by $\bX^{j|=z}$ we refer to an entire dataset in which the value of the $j$-th feature has been set to $z$ for all observations. By $\bX^{\ast j}$, we refer to a modified dataset $\bX$ where the $j$-th feature is randomly permuted and extend this notation to $\bX^{\ast \bj}$ for the case when the set $\bj=\{j_1, ..., j_k\}$ of features is randomly permuted (without breaking the intra-group dependencies).

\subsection{Time-dependent feature effects}

Individual conditional expectation plot~\citep{goldstein2015peeking}, also called the ``What-if?'' plot or ``Ceteris-paribus'' profile~\citep{biecek2021explanatory}, visualizes the local effect of a particular feature on the model's prediction. We define \emph{time-dependent individual conditional expectation} as
\begin{equation}
    \textsc{ice}_t(f, \bx, j, \bz) = \{ f_t(\bx^{j|=z}) \}_{z\in \bz}.
\end{equation}
Specifically, the explanation shows a set of predictions $f$ at time-step $t$ for the observation $\bx$, in which the $j$-th feature spans over a set of its values $\bz$. For example, for a binary feature $j$ we have $\bz = \{0{,} 1\}$ and the explanation can be visualized as two survival functions over a set of time steps $(t_1, \ldots, t_m)$. Intuitively, it answers the local question of ``What would the model predict for observation $\bx$ if the value of feature $j$ was $1$ instead of $0$?''

Partial dependence plot~\citep{friedman2001greedy} aims to identify the global effect of a particular feature on the expected value of predictions. We define the \emph{time-dependent partial dependence plot} as
\begin{equation}
    \textsc{pdp}_t(f, \bX, j, \bz) = \left\{ \operatorname{E}_{\mathcal{X}^{-j}} \left[ f_t(\bX^{j|=z}) \right] \right\}_{z\in \bz},
\end{equation}
where $\mathcal{X}$ represents a $p$-dimensional random variable from which distribution the dataset $\bX$ was collected. The symbol $\mathcal{X}^{-j}$ denotes its modification obtained by removing the $j$-th component, and $\bX^{j|=z}$ is an extension where the $j$-th component's value of $\bX$ is replaced with constant value $z$. In practice, the expected value in $\textsc{pdp}_t$ is estimated as
\begin{equation}
    \widehat{\textsc{pdp}}_t(f, \bX, j, \bz) = \left\{ \frac{1}{n} \sum_{\bx \in \bX}  f_t(\bx^{j|=z}) \right\}_{z\in \bz},
\end{equation}
i.e. an average of local $\textsc{ice}_t$ explanations is calculated over a set of $n$ observations. Note that alternative estimators of global feature effects~\citep{apley2020visualizing,gkolemis2023dale} can be similarly adapted to interpret time-to-event predictions.

\subsection{Time-dependent global feature importance}

One approach to quantify each feature's importance is based on how much predictive power it contributes~\citep{covert2020understanding}. We can measure time-dependent global feature importance as an aggregation of local SurvSHAP(t), which estimates feature attributions $\phi_t(\bx, j)$ with either sampling or kernel approximation (refer to~\citep{krzyzinski2023survshap} for details). We define global \emph{SurvSHAP(t) feature importance} as
\begin{equation}
    \textsc{sfi}_t(f, \bX, j) = \frac{1}{n} \sum_{\bx \in \bX} \mid \phi_t\left( \bx, j \right) \mid,
\end{equation}
i.e. an average feature attribution over a set of observations $\bX$, e.g. a validation set.

Alternatively, permutation feature importance~\citep{breiman2001random,fisher2019all} measures the global influence of a particular feature on the model's performance, e.g. Brier score. Let us denote by $\mathcal{L}_t(f, \bX, \by)$ a performance metric (or a loss function) for a survival model at time-step $t$, where $\by$ is a target feature corresponding to dataset~$\bX$. We define \emph{time-dependent permutation feature importance} as
\begin{equation}\label{eq:pfi}
    \textsc{pfi}_t(f, \bX, j, \mathcal{L}, \by) = \frac{1}{b} \sum_{i=1}^{b} \left( \mathcal{L}_t(f, \bX, \by) - \mathcal{L}_t(f, \bX^{\ast j_i}, \by) \right),
\end{equation}
which effectively measures the impact on the prediction's performance of breaking the dependency between feature $j$ and target feature; aggregated over $b$ permutations of feature $j$. Note that $\textsc{pfi}_t$ should be estimated on the validation set as it requires access to ground truth~$\by$. In practice, we visualize the absolute time-dependent importance $\mid \textsc{pfi}_t(f, \bX, j, \mathcal{L}, \by) \mid$ to be consistent between decreasing and increasing performance metrics, e.g. Brier score vs. AUC. In some applications, it might be useful to consider relative feature importance using $\frac{\mathcal{L}_t(f, \bX^{\ast j}, \by)}{\mathcal{L}_t(f, \bX, \by)}$ instead of $\mathcal{L}_t(f, \bX, \by) - \mathcal{L}_t(f, \bX^{\ast j}, \by)$ in Equation~\ref{eq:pfi}.

In many scenarios, data comprises features that can be meaningfully grouped, e.g. in a multi-modal case where features come from varied modalities. It is then useful to consider grouped feature importance measures~\citep{au2022grouped}. We define \emph{time-dependent grouped permutation feature importance} as
\begin{equation}
    \textsc{gpfi}_t(f, \bX, \bj, \mathcal{L}, \by) = \frac{1}{b} \sum_{i=1}^{b} \left( \mathcal{L}_t(f, \bX, \by) - \mathcal{L}_t(f, \bX^{\ast \bj}, \by) \right).
\end{equation}
That is, we analyze the influence of breaking the dependency of features from a particular group $\bj$ and all other features while preserving the intra-group relationships. We acknowledge that grouping features in Shapley-based explanations is possible~\citep{au2022grouped}, but leave that as future work.

\section{Materials: data and machine learning models}\label{sec:materials}

We show the applicability of time-dependent explanations in two applications. Section~\ref{sec:materials-tlos} describes in detail the process of ML analysis from data acquisition through model selection towards explaining time-to-event LoS predictions. Section~\ref{sec:materials-tcga} focuses more on a large-scale approach to biomarker discovery using the introduced explanation methods. We will discuss the obtained results in Sections~\ref{sec:results-tlos}~\&~\ref{sec:results-tcga}, respectively.

\subsection{Bias in predicting hospital LoS using X-ray images}\label{sec:materials-tlos}

IML methods for time-to-event prediction can be valuable for explaining hospital LoS to humans. We consider a setting where LoS is predicted using time-to-event survival models instead of single-value time estimation with regression models or time-span classification. Besides giving a more holistic prediction, survival analysis naturally allows for censored observations in data, e.g. a patient was discharged from one hospital and moved to another with further information missing. 

\paragraph{The \textsc{tlos} dataset} A unique multi-modal dataset used in this study is created based on image, text and tabular data of 1235 patients from the University Clinical Centre, Medical University of Warsaw, Poland. It allows us to answer the question of interest: \textbf{To what extent can the patient's LoS in a hospital be predicted using an X-ray image?} The \emph{target feature} is the time between the patient's radiological examination and hospital discharge (in days, $min=1$, $median=7$, $mean=13.73$, $max=330$). Due to the high skewness of the time distribution, we model the logarithm of time in practice. About 20\% of outcomes are right-censored, e.g. due to death. The dataset includes 749 (61\%) male and 486 (39\%) female patients of rather evenly distributed age (in full years, $min=0$, $mean=38$, $median=37$, $max=90$). 

\begin{figure}
    \centering
    \includegraphics[width=\columnwidth]{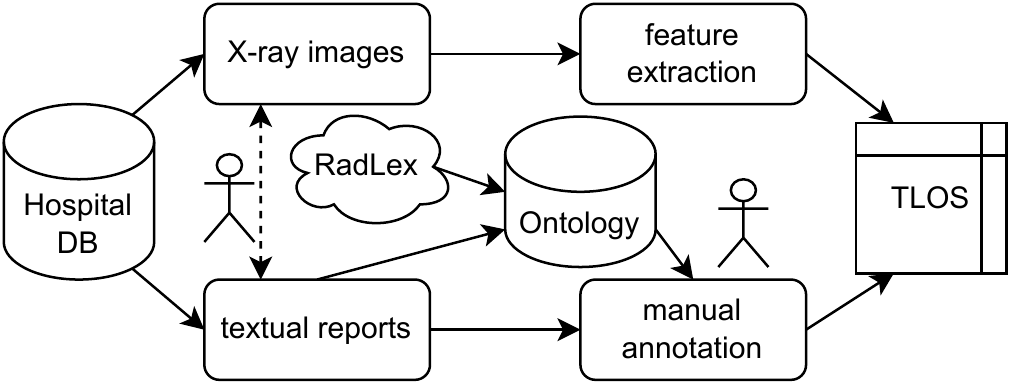}
    \caption{Schematic workflow of creating the \textsc{tlos} dataset.}
    \label{fig:dataset}
\end{figure}

Figure~\ref{fig:dataset} shows a high-level workflow of creating the \textsc{tlos} dataset. Each radiologic exam consists of an X-ray image with a written report stating observable features, e.g. pathological signs, lung lesions and pleural abnormalities, but also the occurrence of medical devices on the image, e.g. tubing and electrocardiographic leads. We obtained X-ray images of high quality in the raw DICOM format, which were then converted to PNG and resized to 420x420 before further processing. We manually annotate each textual report into 17 interpretable binary features informing whether the pathology occurs or is absent. Note that we sampled patients at random and capped their quantity after reaching the reasonable capacity of human annotators. Moreover, we automatically extract 76 numerical features from the image using the \texttt{pyradiomics} tool~\cite{van2017computational}. It computes various statistics based on an image and a lung segmentation mask, e.g. various aggregations of the grey-level co-occurrence matrix. A detailed description of algorithm-extracted features from \texttt{pyradiomics} is available at \url{https://pyradiomics.readthedocs.io/en/v3.0.1/features.html}. A pretrained CE-Net~\cite{gu2019net} model was used to obtain lung segmentation masks inputted to \texttt{pyradiomics}. We treat it as a reasonable baseline approach while acknowledging that segmentation errors will inevitably contribute to errors in algorithm-extracted features. The described procedure leads to obtaining four feature sets referred to as:
\begin{itemize}
    \item \emph{baseline} (number of features: $d=2$) -- includes the patient's age and sex,
    \item \emph{human-annotated} ($d=2+17$) -- includes \emph{baseline} and pathology occurrences,
    \item \emph{algorithm-extracted} ($d=2+76$) -- includes \emph{baseline} and radiomics features, 
    \item \emph{all features} ($d=2+17+76$).
\end{itemize}

\paragraph{Details of human annotation} The original raw dataset included X-ray examinations, i.e. images with textual radiology reports, which we anonymized extensively. First, we developed a project-specific ontology of chest pathologies that can be observed on X-ray images. It compiles information from the relevant radiology literature~\citep{hansell2008fleischner}, existing ontologies like RadLex~\citep{radlex}, and popular X-ray databases available online (CheXpert~\citep{irvin2019chexpert}, MIMIC~\citep{johnson2023mimic}, VinDr-CXR~\citep{nguyen2022vindr}). Second, we identified problems in our reports related to, e.g. ambiguous interpretations of phrases in the radiological reports, differences in the quality and length of descriptions prepared by different physicians and for different types of examinations (like follow-up or comparison to previous examinations). We fixed them by updating the ontology in accordance with the domain knowledge. As a result, a set of 35 classes was obtained. Two board-certified radiologists annotated whether a given class occurs~(1) or not~(0) in a textual report from the X-ray examination, which was consistent. After annotation, we chose the 17 most common features, i.e. with more than 3\% occurrence among patients, for further analysis in this study. Although we are yet unable to share X-ray images and textual reports due to privacy concerns, we share the preprocessed \textsc{tlos} dataset with further documentation and code to reproduce our results (see Data and code availability). 

\paragraph{Models}  We first use all features to compare various ML survival models in predicting LoS using X-ray images and then evaluate the impact of particular feature sets on the best models' predictive performance. We perform a comprehensive benchmark of relevant learning algorithms available in the \texttt{mlr3proba} toolbox~\cite{sonabend2021mlr3proba}: a decision tree (CTree), gradient boosting decision trees (GBDT), two implementations of random survival forest (Ranger \& RF-SRC), Cox proportional hazards (CoxPH), and two implementations of neural networks (DeepSurv \& DeepHit). For details of the models' hyperparameters refer to~\ref{app:hyperparameters}.  We use 10 repeats of 10-fold cross-validation and assess the predictive performance of survival models with two metrics: C-index, where higher means better performance and the baseline value of a random model equals 0.5, and integrated Brier score (IBS), where lower is better and the baseline value of a random model equals 0.25. The evaluation protocol mimics a benchmark of survival prediction methods described in~\cite{herrman2021benchmark}. Moreover, we report integrated AUC (IAUC) for the models, for which it is implemented (CoxPH and GBDT).

\subsection{Explainable multi-omics for cancer survival prediction}\label{sec:materials-tcga}

To showcase the versatility of the introduced methods for explaining ML survival models, in Section~\ref{sec:results-tcga}, we apply them to interpret predictions of cancer survival. Our goal is to analyse the importance of features grouped in varied modalities, which is valuable for model developers and medical experts aiming to find the best type of multi-omics biomarkers. 

\paragraph{The TCGA benchmark} To this end, we rely on a large-scale multi-modal benchmark consisting of standardised datasets derived from The Cancer Genome Atlas (TCGA) database~\citep{herrmann2021largescale}. Each dataset considers a different cancer survival target feature, relevant clinical features and 4 types of high-dimensional molecular features, e.g. miRNA expressions. The target feature is right-censored. Refer to~\citep[][table 2]{herrmann2021largescale} for a detailed summary of the datasets. Our goal is to create lower-dimensional multi-modal datasets appropriate for the purpose of IML. Thus, we follow the work of~\citep{bommert2022benchmark}, where a benchmark of 14 feature selection methods was conducted on 11 out of 18 datasets with more than 50 events. For each dataset, we apply the best-performing feature selection method, i.e. variance filter~\citep{bommert2022benchmark}, to choose the 5 most relevant features per each modality (including clinical features). This procedure results in an approachable benchmark of 11 datasets, each with 25 multi-modal features. 

\paragraph{Models} We use RSF on all datasets as a robust baseline model. Models are evaluated with the IBS metric based on 10-fold cross-validation. Importance scores are aggregated in feature groups representing various omics modalities and averaged over test sets in cross-validation folds. Feature effect is estimated based on a separate RSF trained on the whole dataset, which aims to mimic the workflow of testing the significance of parameters in a CoxPH model.

\section{Results}\label{sec:results}

\subsection{Bias in predicting hospital LoS using X-ray images}\label{sec:results-tlos}

\paragraph{Model performance} Table \ref{tab:model_comparison} presents the results where models are sorted by an average C-index. First, note that most DeepHit models did not converge and provide random predictions; thus, DeepHit is removed from the comparison. We observe that, on average, the best algorithm is a gradient boosting decision tree (0.668 C-index, 0.117 IBS) with a random survival forest in second. Interestingly, the interpretable and widely-used CoxPH model performs worse (0.645 C-index, 0.127 IBS). Overall, neural networks have a hard time learning meaningful models. Comparing the raw performance values with results from related work leads to the conclusion that \emph{predicting the patient's LoS from an X-ray image is indeed possible}, but challenging. 

\begin{table}
    \centering
    \begin{tabular}{l|ccc}
        \toprule
        \textbf{Model} & \textbf{C-index} $\uparrow$ & \textbf{IBS} $\downarrow$ & \textbf{IAUC} $\uparrow$ \\
        \midrule
        DeepSurv & $.628_{\pm .036}$ & $.127_{\pm .016}$ & -- \\
        Ranger & $.639_{\pm .027}$ & $.140_{\pm .010}$ & -- \\
        CTree & $.641_{\pm .030}$ & $.127_{\pm .015}$ & -- \\
        CoxPH & $.645_{\pm .029}$ & $.127_{\pm .013}$ & $.678_{\pm .056}$ \\
        RF-SRC & $.651_{\pm .030}$ & $.119_{\pm .013}$ & -- \\
        GBDT & $.668_{\pm .030}$ & $.117_{\pm .014}$ & $.715_{\pm .061}$ \\
        \bottomrule
    \end{tabular}
    \caption{Benchmark of ML survival models predicting LoS using all features from X-ray images sorted by C-index. Most of the DeepHit models did not converge and provide random predictions; thus, we removed it from the comparison. Based on 10 repeats of 10-fold cross-validation, the GBDT algorithm performs best on average, i.e. achieves 0.668 C-index and 0.117 IBS. For broader context, we report IAUC for CoxPH and GBDT, which are the only algorithms to implement this metric. Detailed results are in~\ref{app:performance}.}
    \label{tab:model_comparison}
\end{table}

\begin{table}
    \centering
    \begin{tabular}{ll|ccc}
        \toprule
        \textbf{Model} & \textbf{Features} & \textbf{C-index} $\uparrow$ & \textbf{IBS} $\downarrow$ & \textbf{IAUC} $\uparrow$ \\
        \midrule
    \multirow{4}{*}{CoxPH} & baseline & $.567_{\pm .036}$ & $.129_{\pm .013}$ & $.570_{\pm .060}$ \\
        & AE & $.639_{\pm .026}$ & $.127_{\pm .014}$ & $.677_{\pm .051}$ \\
        & HA & $.642_{\pm .028}$ & $.120_{\pm .014}$ & $.676_{\pm .056}$ \\
        & all feat. & $.645_{\pm .029}$ & $.127_{\pm .013}$ & $.678_{\pm .056}$ \\
        \midrule
    \multirow{4}{*}{GBDT} & baseline & $.582_{\pm .037}$ & $.126_{\pm .013}$ & $.599_{\pm .064}$ \\
        & AE & $.649_{\pm .029}$ & $.121_{\pm .014}$ & $.702_{\pm .057}$ \\
        & HA & $.652_{\pm .026}$ & $.119_{\pm .014}$ & $.686_{\pm .054}$ \\
        & all feat. & $.668_{\pm .030}$ & $.117_{\pm .014}$ & $.715_{\pm .061}$ \\
        \bottomrule
    \end{tabular}
    \caption{Benchmark of four feature sets -- baseline, algorithm-extracted (AE), human-annotated (HA), and all features -- in predicting LoS using GBDT black-box and CoxPH model (white-box). We report average and standard deviation of performance based on 10 repeats of 10-fold cross-validation. Detailed results with \emph{p}-values are in~\ref{app:performance}.}
    \label{tab:blackbox_whitebox_comparison}
\end{table}

\paragraph{Feature performance} We first aim to tackle the performance--interpretability tradeoff in ML for medicine~\cite{rudin2019stop}. Based on the benchmark results reported in Table~\ref{tab:model_comparison}, we choose the best black-box algorithm (GBDT) to compare with CoxPH -- the widely-adopted interpretable approach to time-to-event analysis. Note that one can consider the human-annotated features as interpretable and algorithm-extracted features as a black-box approach, i.e. training the CoxPH model on algorithm-extracted features is not necessarily a white-box model. We test the two algorithms on four feature sets using the same cross-validation scheme and performance metrics as the previous benchmark. Table~\ref{tab:blackbox_whitebox_comparison} presents the results where feature sets are sorted by an average C-index. We observe that both algorithm-extracted and human-annotated features include valuable information for predicting LoS. The only significant difference (on average) between the two sets is for the CoxPH algorithm evaluated with IBS. For GBDT, using all features results in the best performance, while for CoxPH, increasing the number of features leads to the same or worse performance due to the curse of dimensionality. In summary, the best-performing interpretable algorithm is CoxPH trained on human-annotated features (0.642 C-index, 0.120 IBS, 0.676 IAUC), and the black-box approach is GBDT trained on all features (0.668 C-index, 0.117 IBS, 0.715 AUC). 
The difference in C-index and IAUC is significant (\emph{p}-value $< 0.001$). 
Although this difference may be neglectable in reality, CoxPH is limited by the number of features, which now rapidly increases in medical applications, e.g. in radiology where tools for feature extractions become more prevalent~\citep{chaou2017predicting}. Moreover, annotating images by humans is costly, and GBDT remains more efficient with algorithm-extracted features.

\paragraph{Explaining length of stay predictions to humans} A classic approach to the explanatory analysis of time-to-event models involves analysing the significance of the CoxPH model's coefficients. For a broader context, we use all observations in data to fit CoxPH models to the four feature sets. We report features with significant coefficients in~\ref{app:significance}, effectively serving as a list of features important to predicting LoS. \textbf{The main limitation of this approach is explaining only a particular learning algorithm that performs worse in the predictive task, as in our case.} Therefore, we propose to use model-agnostic explanations to interpret the predictions of any black-box survival model predicting LoS in general.  We extend the explanatory model analysis framework~\citep{baniecki2023grammar} to include time-dependent explanations. For a concrete example, we use all observations in data and fit a GBDT model to the human-annotated feature set. 

\begin{figure}[!h]
    \centering
    \includegraphics[width=\columnwidth]{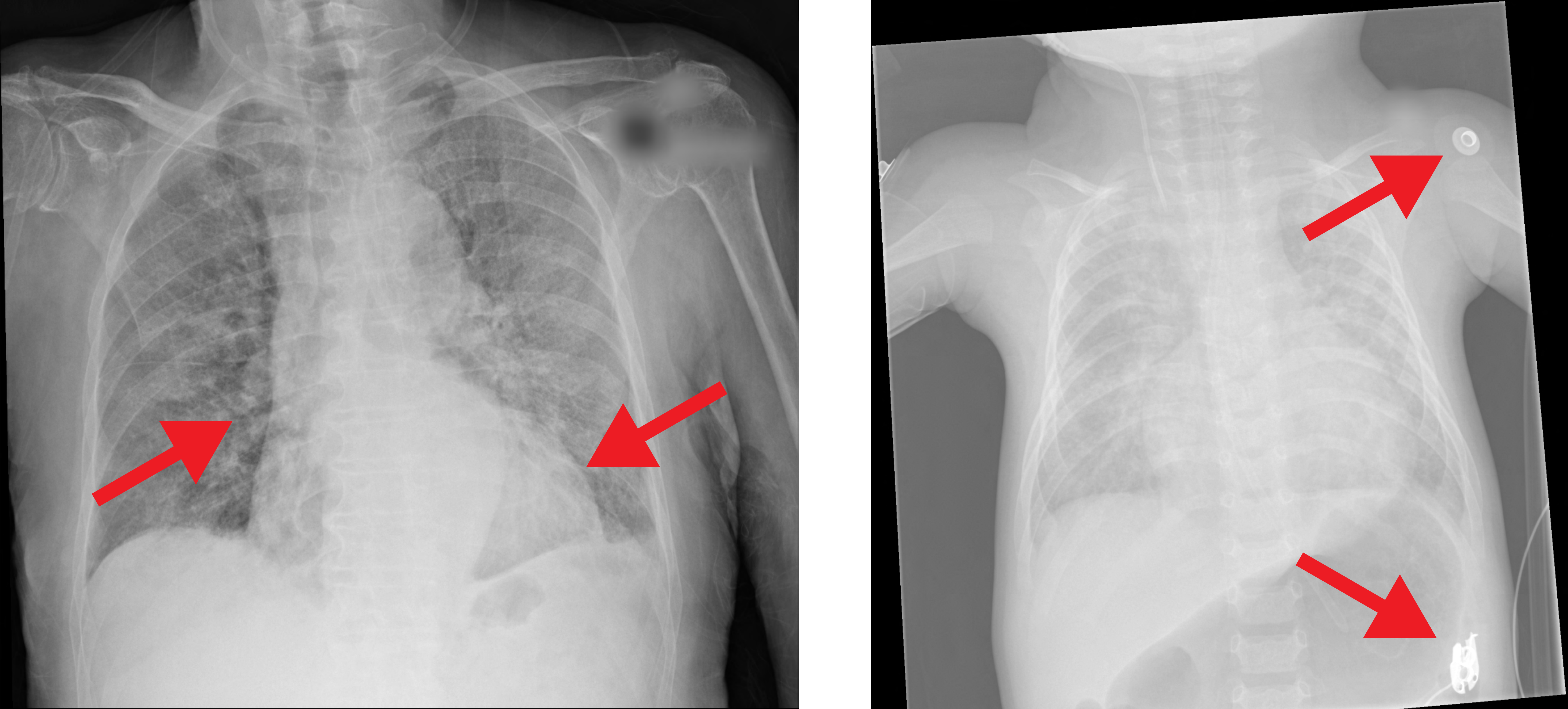}
    \caption{Exemplary X-ray images of (\textbf{left}) lung disease in an adult patient and (\textbf{right}) healthy children's lungs with visible medical devices.}
    \label{fig:examples}
\end{figure}

\begin{figure*}
    \centering
    \includegraphics[width=\textwidth]{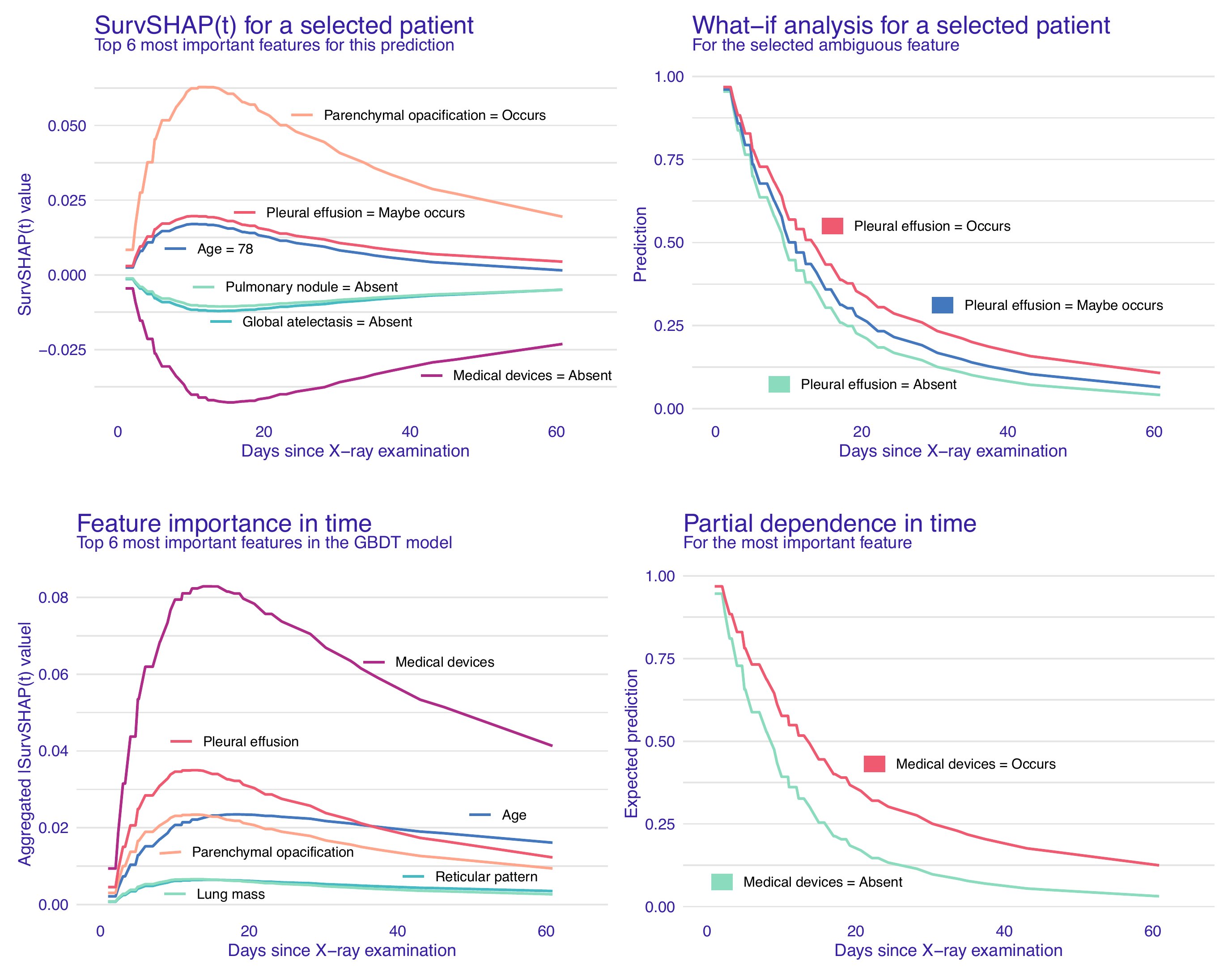}
    \caption{Complementary time-dependent explanations of the GBDT model trained on age, sex, and human-annotated radiomics features. \textbf{Top left}: SurvSHAP(t) local explanation for a selected patient informing about the 6 most important features and their effect on the predicted LoS on each day since X-ray examination. \textbf{Top right}: What-if analysis for the same patient and the selected ambiguous feature, informing about how the predicted LoS would change upon the change in feature value. \textbf{Bottom left}: Feature importance global explanation based on aggregated SurvSHAP(t) values for a subset of patients. It informs about the 6 most important features overall. \textbf{Bottom right}: Partial dependence global explanation for the most important feature informing about its effect on the predicted LoS.}
    \label{fig:blackbox_explanations}
\end{figure*}

Figure~\ref{fig:examples} shows two X-ray image samples: one with lung diseases and another with healthy lungs and medical devices. Figure~\ref{fig:blackbox_explanations} presents four complementary local and global explanations, providing a multi-faceted understanding of the black-box model.  We first interpret the particular prediction for a 78-year-old male patient with parenchymal opacification in the lungs and possibly also pleural effusion. SurvSHAP(t) attributes high importance to the occurrence of parenchymal opacification, but also the absence of medical devices on the chest X-ray. \textbf{In fact, the latter decreases the predicted probability of longer LoS as X-rays with observable medical devices usually indicate a severe patient condition.} This image feature is a potentially harmful bias in data that later propagates to a predictive model. Next, we perform a What-if analysis for the ambiguous pleural effusion feature to explain the uncertainty in LoS prediction conditioned on this feature. Over the 60 days since the X-ray examination, the probability of staying in a hospital is increased by 0.125 when a pleural effusion occurred (for this patient).

The bottom of Figure~\ref{fig:blackbox_explanations} presents global explanations of the model's behaviour: feature importance and effects, also referred to as partial dependence. We obtain time-dependent feature importance by aggregating absolute SurvSHAP(t) values for a representative subset of patients. The visualization indicates that the model finds the occurrence of medical devices as a proxy for LoS, which may be correlated with the patient's condition. It is an evident bias in data, which can be an enormous risk for model deployments in medicine and healthcare (see e.g. work on racial bias in AI recognition of medical images~\citep{gichoya2022ai}). Other important radiomics features include pleural effusion, age, parenchymal opacification, cardiac silhouette enlargement and lung mass. For each of these pathologies, a partial dependence plot explains its aggregated effect on the LoS prediction. Specifically, when medical devices occur on an image, the probability of staying in a hospital 20 days after the X-ray examination is increased by 0.25.

As illustrated here, incorporating time-dependent explanations of ML models predicting LoS into the existing decision support systems can provide useful information for physicians. The presented approach is general and widely applicable to other time-to-event medical use cases.

\subsection{Explainable multi-omics for cancer survival prediction} \label{sec:results-tcga}

\begin{figure*}[t]
    \centering
    \includegraphics[width=\textwidth]{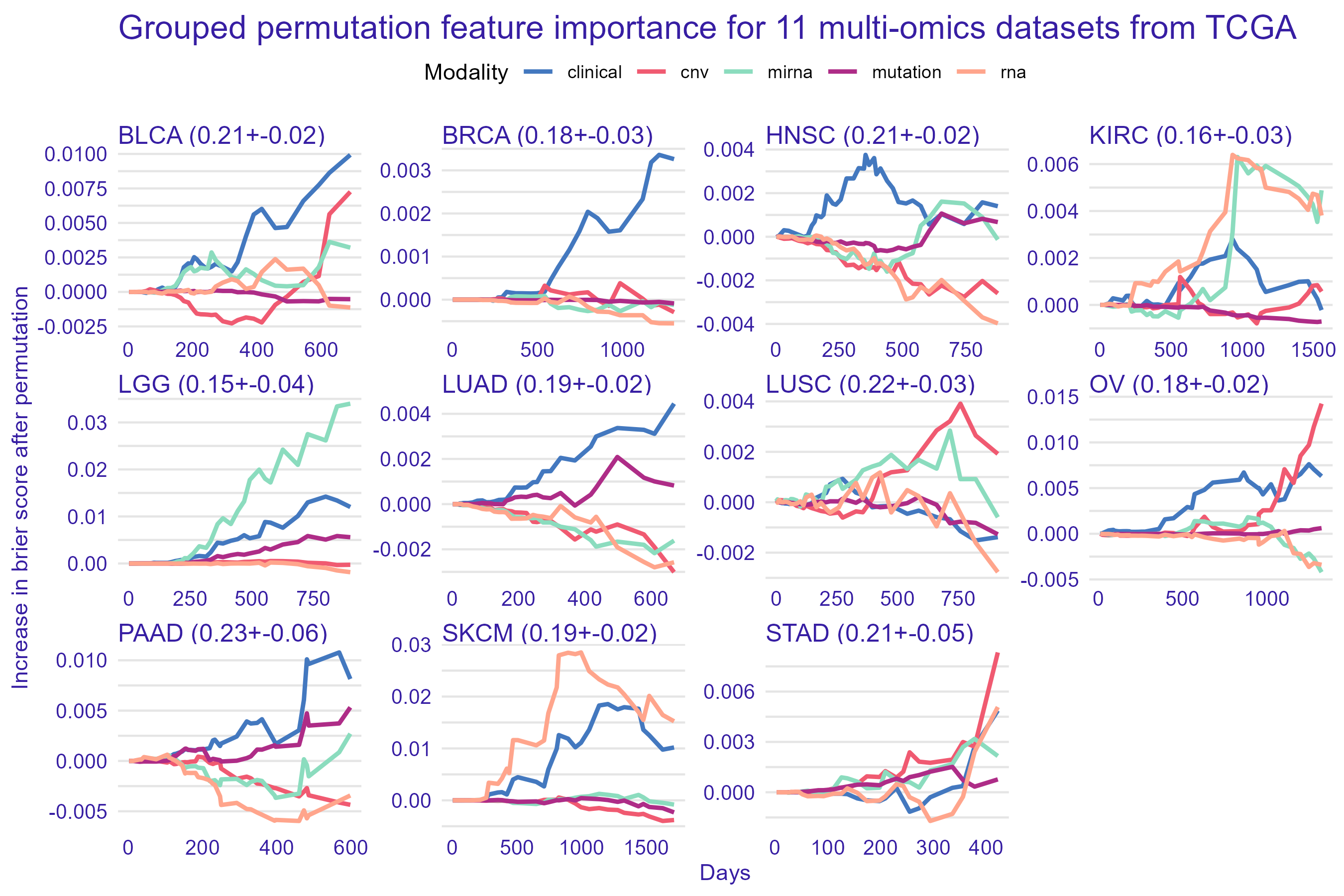}
    \caption{Time-dependent permutation importance of feature groups representing different omics modalities. IBS metric values are in brackets. Each subplot comprises distinct results from a cross-validation of a random survival forest predicting the survival of a particular cancer type from TCGA. Positive importance values over time indicate a high influence of features on the accuracy of time-to-event prediction. We observe that clinical features are the most important for predicting BLCA, BRCA, HNSC, LUAD, OV, and PAAD; CNV features influence the prediction in LUSC and OV; miRNA features are important in KIRC, LGG and LUSC; while RNA features are useful in predicting KIRC and SKCM. In some tasks, particular modalities add noise to the modelling process, as shown by negative importance values over time. For example, a random survival forest predicting PAAD performs the worst on average, practically random. It is indicated by the integrated Brier score of $0.23\pm0.06$, as well as in the time-dependent explanation showing most of the features as unimportant noise.}
    \label{fig:tcga_gpfi_cv}
\end{figure*}

\begin{figure}[!ht]
    \centering
    \includegraphics[width=\columnwidth]{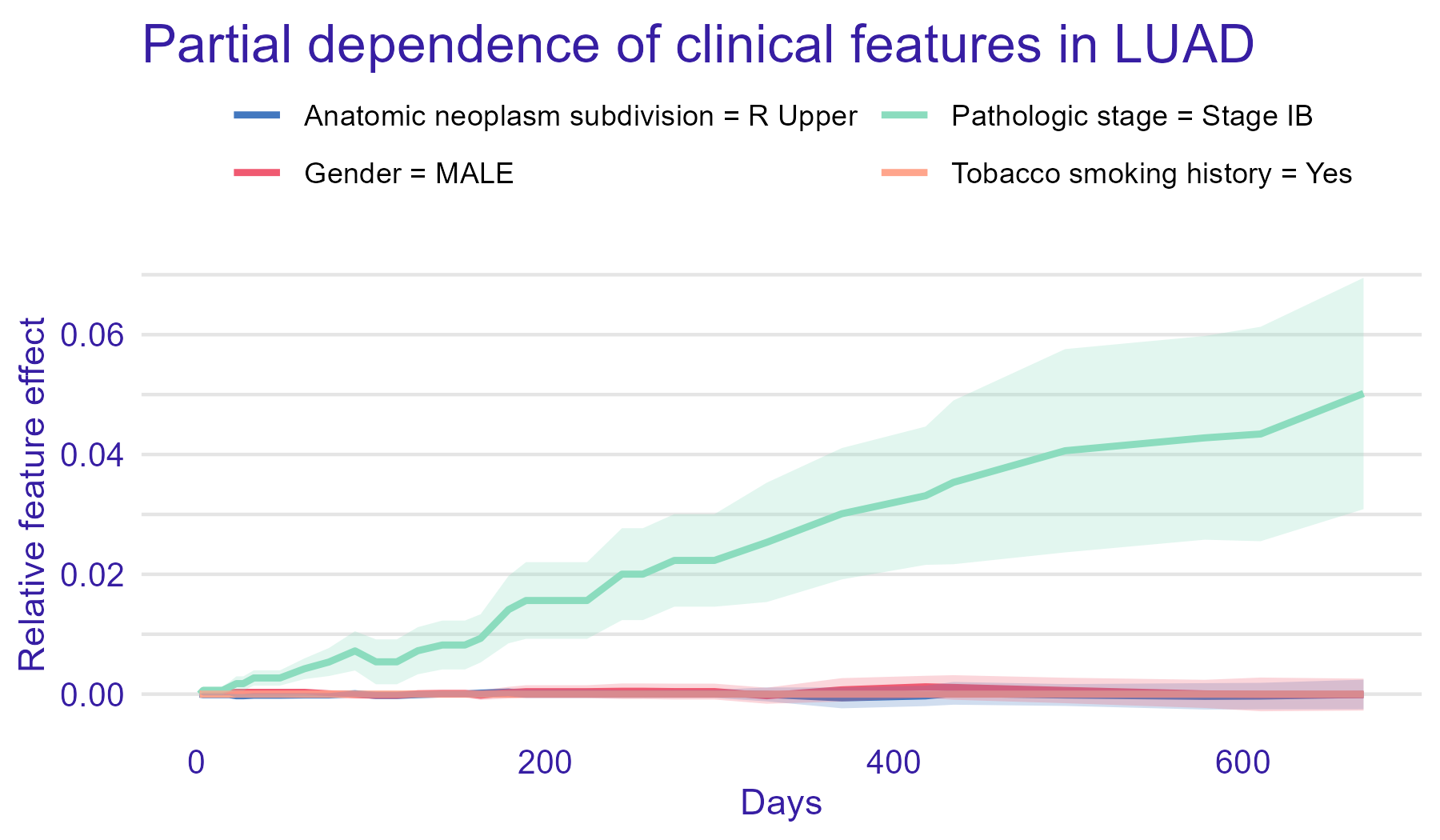}
    \caption{Relative effects of binary clinical features in predicting the survival of lung adenocarcinoma (LUAD). We visualise the relative difference between effects for feature values 1 and 0, where the ribbon denotes the standard deviation of the estimator. Early pathologic stage (1B) of cancer increases time-to-event predicted by random survival forest. Other clinical features seem unimportant.}
    \label{fig:tcga_pdp_luad}
\end{figure}

\begin{figure}[!ht]
    \centering
    \includegraphics[width=0.95\columnwidth]{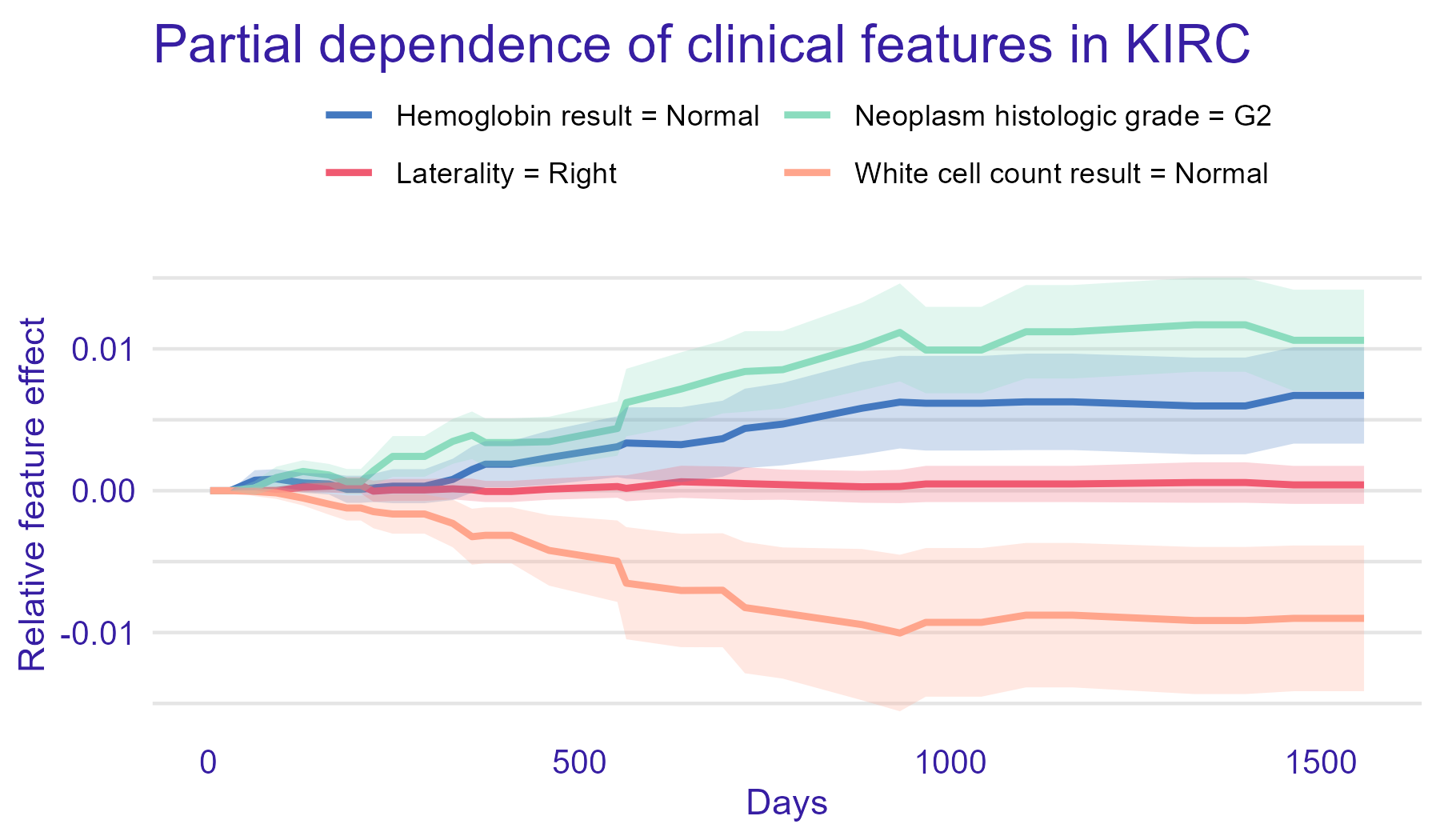}
    \caption{Relative effects of binary clinical features in predicting the survival of kidney renal clear (KIRC) cell carcinoma. We visualise the relative difference between effects for feature values 1 and 0, including standard deviation.
    }
    \label{fig:tcga_pdp_kirc}
\end{figure}

\paragraph{Time-dependent feature importance} Figure~\ref{fig:tcga_gpfi_cv} shows explanations of time-dependent grouped permutation feature importance for RSF trained on each dataset from TCGA. Only in some cases a random survival forest achieves satisfactory predictive performance as measured with the IBS metric, e.g. LGG and KIRC. Explaining the importance of feature modalities can provide insights into the challenge of accurate survival analysis. It guides stakeholders in planning to gather more data from a particular modality or improve its quality. Note how the ranking of feature importance changes over time. We observe that, on average, clinical and miRNA features are the most influential to the models' behaviour as judged by the area under the curve. Which features are significant?

\paragraph{Time-dependent feature effects} Partial dependence plots extend and complement the analysis of permutation feature importance. To better understand a specific predictive task, we should analyse the effects of each feature in detail. Figure~\ref{fig:tcga_pdp_luad} shows time-dependent partial dependence of clinical features on RSF predicting survival of lung adenocarcinoma (LUAD). In this simple example, cancer's early pathologic stage (1B) increases (on average) the predicted time-to-event. Moreover, the standard deviation of the estimated explanation provides intuition about its significance. Other clinical features seem unimportant to the model. Time-dependent explanations of ML models can complement classical survival analysis methods in knowledge discovery. An analogous explanation of clinical features in RSF predicting survival of kidney renal clear (KIRC) cell carcinoma is shown in Figure~\ref{fig:tcga_pdp_kirc}. Perhaps counterintuitively, a normal level of white cell count has a negative impact (on average) on the predicted survival function. Is this a bias in data or an undesirable relationship encoded in the black-box predictive model? An explanation can help to debug ML algorithms and align them with domain knowledge.

\section{Discussion} \label{sec:discussion}

\subsection{Towards time-dependent global explanations}

In many medical and healthcare predictive tasks, typical post-hoc interpretability methods were not enough to faithfully explain survival models~\cite{kovalev2020survlime,wang2021counterfactual,rad2022extracting,utkin2022survnam,stone2022systematic}. In \citep{krzyzinski2023survshap}, we proposed the notion of \emph{time-dependent} explanations as a comprehensive visualization tool for data analysis and model debugging. In this paper, we formally introduce the theory behind global time-dependent explanations and show two concrete use cases in which they are useful. First, our approach makes it easy to relate time-dependent predictive performance of survival models to their explanations. Feature importance can be integrated in time to obtain a simpler view similarly to how Brier score is conventionally integrated. Second, physicians obtain access to full information about how specific features and modalities impact the predicted survival time of patients. Our results clearly show different feature rankings and effects at different time points indicating the need for an in-depth analysis of machine learning survival models.

\subsection{Limitations}

\paragraph{Time-dependent explanations} We foresee several limitations of the introduced methods related to general limitations of post-hoc interpretability methods~\citep{molnar2020general}. First, feature effects and feature importance measures implicitly assume that features are independent, which is violated in practice where out-of-distribution observations influence explanation estimation~\citep{apley2020visualizing}. For example, \textsc{ice} fixes the value of a particular feature and \textsc{pfi} ``removes'' the influence of a particular feature by perturbation~\citep{molnar2023model}. In such cases, the conclusions drawn may be misleading or simply wrong~\citep{aas2021explaining}. It is important to acknowledge that there are no \emph{perfect} explanations and they must be interpreted with care. Crucially, different post-hoc interpretability methods may diverge in an explanation of the predictions~\citep{turbe2023evaluation}. Moreover, interpretability methods are vulnerable to adversarial attacks~\citep{baniecki2024advxai,noppel2024explainable}, e.g. minimal perturbations in input data and model parameters that lead to misleading explanations, which should be especially considered in practical, high-risk scenarios. Specifically in case of the proposed time-dependent explanations, important limitations arise from a perceptual (human) viewpoint. While being able to observe the dependence on time is desirable, it introduces an additional dimension for visualization that might lead to information overload~\citep{poursabzi2021manipulating,baniecki2023grammar}. Therefore, future work is needed regarding the perception of potentially illegible plots showing explanations with many curves and for features with large number of unique values.

\paragraph{Dataset collection and annotation} The biases encoded in data and machine learning algorithms are an important concern, especially in medicine~\citep{irvin2019chexpert,nguyen2022vindr,johnson2023mimic}. In Section~\ref{sec:materials-tlos}, we describe the novel dataset developed for the purpose of our work on time-dependent explanations. Its collection is certainly biased with respect to geographic localization (Poland) and type of patients (lung screening). We include patients' age and sex to account for the latter. As for data annotation, we minimize risks related to human bias by having two radiologists annotate the data, and thus minimize the interobserver error. We contribute an openly available dataset of over one thousand patients, which is rather substantial in the domain of LoS prediction. In general, the process of annotating medical data involves human expertise, is rather costly, and not easily scalable~\citep{irvin2019chexpert,nguyen2022vindr,johnson2023mimic}.

\section{Conclusion} \label{sec:conclusion}

We formally introduce time-dependent global feature importance and feature effects to extend the interpretability toolbox for time-to-event prediction. Experiments with multi-modal medical data demonstrate significant added value when performing a comprehensive explanatory analysis of ML models. Notably, post-hoc explanations allow for finding biases in AI systems predicting hospital LoS, and evaluating cancer survival models beyond performance to include the importance of multi-omics feature groups. We hope that the contributed data and code resources facilitate future work in the emerging research direction of explainable survival analysis. 

\paragraph{Future work} It will be valuable to propose interactive ways of visualizing time-dependent explanations for diverse stakeholders from the medical domain~\citep{baniecki2023grammar}. Our initial proposal borrows from the well-established visualization of survival curves, but it can be further improved based on human feedback~\citep{combi2022manifesto}. We need to evaluate the introduced methods in debugging machine learning models developed to analyse data in clinical settings and at the point-of-care. From the computational perspective, one can work on improving the efficiency and accuracy of estimating explanations for complex data~\citep{aas2021explaining} and models~\citep{chen2023algorithms}.

\section*{Declaration of competing interest}
The authors have no known conflicts of interest.

\section*{Acknowledgements}
This work was financially supported by the Polish National Center for Research and Development grant number INFOSTRATEG I/0022/2021-00, and carried out with the support of the Laboratory of Bioinformatics and Computational Genomics and the High Performance Computing Center of the Faculty of Mathematics and Information Science, Warsaw University of Technology.

\section*{Data and code availability} \label{sec:dataandcode}
We publicly share the \textsc{tlos} dataset and contribute the implementations of time-dependent \textsc{ice}, \textsc{pdp}, and \textsc{pfi} in Python at \url{https://github.com/mi2datalab/interpret-time-to-event}. Moreover, \textsc{sfi} is implemented in Python at \url{https://github.com/mi2datalab/survshap}, and time-dependent explanations are implemented in the \texttt{survex} R package~\citep{spytek2023survex}. Datasets comprising the TCGA benchmark can be freely accessed from OpenML~\citep{bischl2021openml}.

\appendix
\section{Hyperparameters of ML models}\label{app:hyperparameters}

See Table~\ref{tab:hyperparameters}.

\begin{table}[ht]
    \centering
    \caption{Non-default hyperparameters of machine learning models evaluated in the benchmark.}
    \label{tab:hyperparameters}
    \vspace{1em}
    \begin{tabular}{llr}
        \toprule
        \textbf{Model} & \textbf{Parameter} & \textbf{Value} \\
        \midrule
        RF-SRC & splitrule & bs.gradient \\
        \midrule
        \multirow{2}{2cm}{GBDT} & nu & 0.01 \\
         & mstop & 2000 \\
        \midrule
        \multirow{5}{2cm}{DeepSurv} & lr & 0.1 \\
         & epochs & 2000 \\
         & optimizer & adadelta \\
         & neurons & 8 \\
         & batch\_size & 64 \\
        \midrule
        \multirow{5}{2cm}{DeepHit} & lr & 1 \\
         & epochs & 2000 \\
         & optimizer & adadelta \\
         & neurons & 8 \\
         & batch\_size & 64 \\
        \bottomrule
    \end{tabular}
\end{table}

\section{Performance of ML models}\label{app:performance}

\begin{figure*}[ht]
    \centering
    \includegraphics[width=\textwidth]{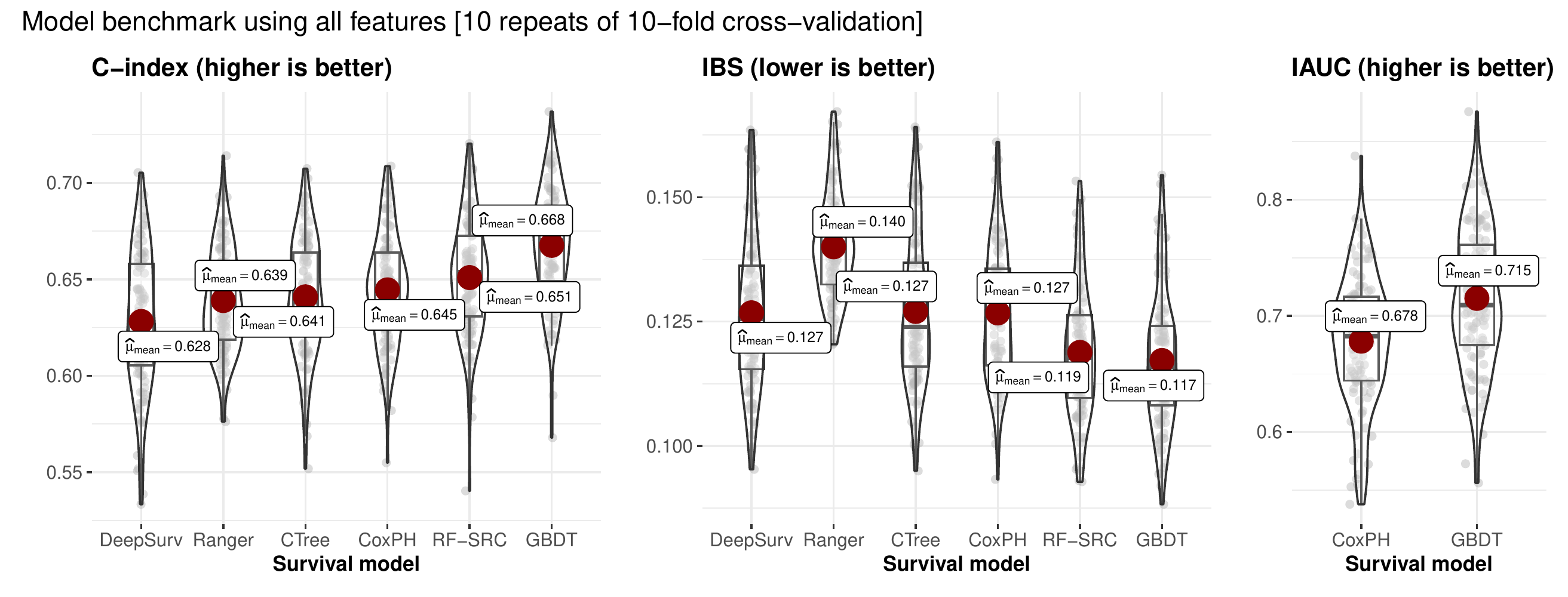}
    \caption{Benchmark of ML survival models predicting LoS using features from X-ray images. Most of the DeepHit models did not converge and provide random predictions; thus, DeepHit is removed from the comparison. Based on 10 repeats of 10-fold cross-validation, the GBDT algorithm performs best on average, i.e. achieves 0.668 C-index and 0.117 IBS. For broader context, we report IAUC for CoxPH and GBDT, which are the only algorithms to implement this metric. In contrast to Figure~\ref{fig:blackbox_whitebox_comparison}, we omit reporting \emph{p}-values for significant differences as there are too many.}
    \label{fig:model_comparison}
\end{figure*}

\begin{figure*}[ht]
    \includegraphics[width=\textwidth]{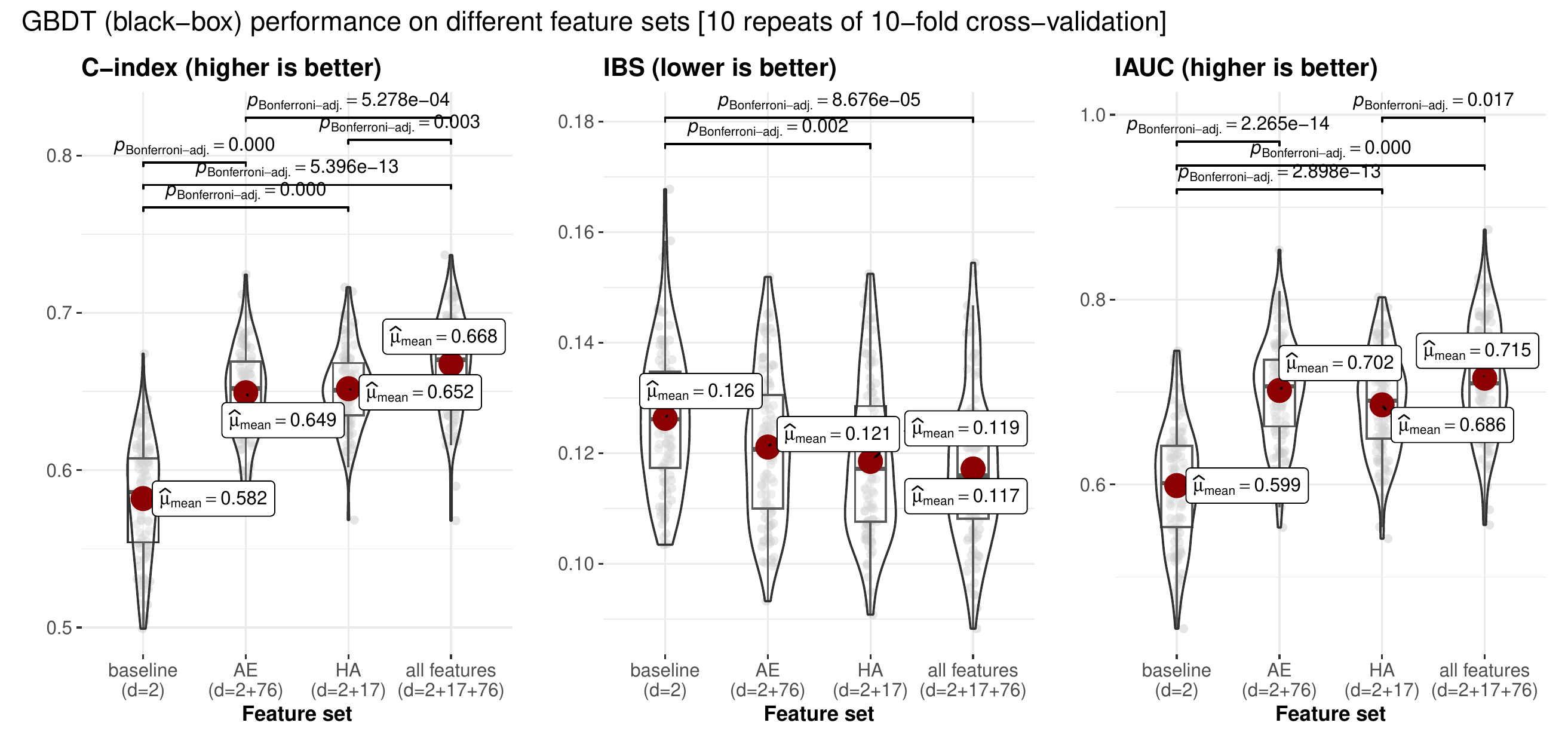}
    \includegraphics[width=\textwidth]{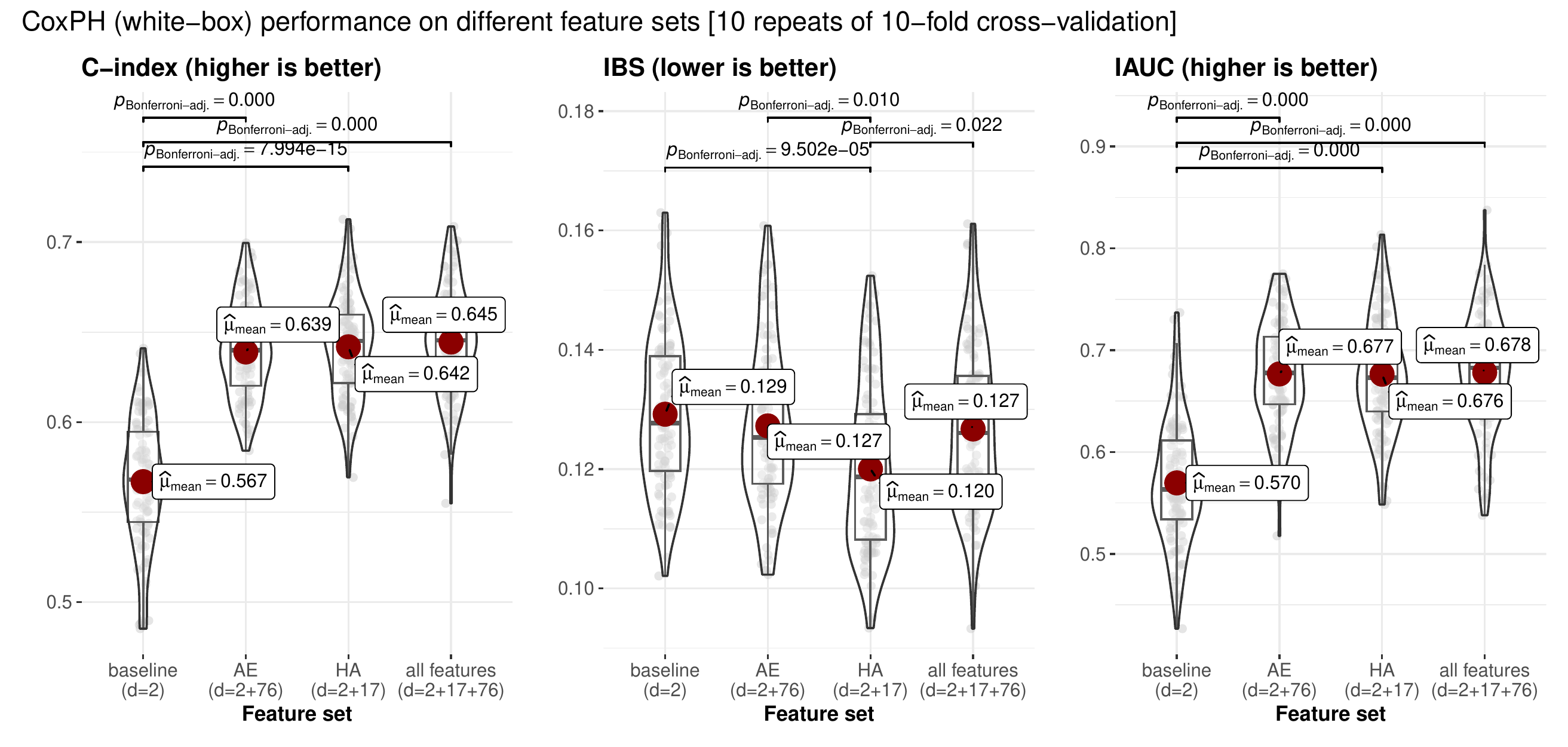}
    \caption{Benchmark of four feature sets -- baseline, algorithm-extracted (AE) and human-annotated (HA), all features -- in predicting LoS using GBDT black-box and CoxPH white-box model. The \emph{p}-values mark significant differences between the average results.}
    \label{fig:blackbox_whitebox_comparison}
\end{figure*}

See Figures~\ref{fig:model_comparison}~\&~\ref{fig:blackbox_whitebox_comparison}.

\section{Significance of features in CPH models}\label{app:significance}

See Table~\ref{tab:coxph}.

\begin{table*}[ht]
    \centering
    \caption{Coefficients of only significant features (\emph{p}-value $<0.05$) in the CoxPH models fitted to all observations and the four feature sets. For example, the CoxPH model fitted on the age and sex of 1235 patients relies on age, not sex, to make the prediction. A detailed description of algorithm-extracted features with their acronyms is available at \url{https://pyradiomics.readthedocs.io/en/v3.0.1/features.html}.} 
    \label{tab:coxph}
    \vspace{1em}
    \begin{tabular}{llwr{2cm}wr{1.5cm}}
      \toprule
    \textbf{Feature set} & \textbf{Feature name} & \textbf{Estimate} & \emph{p}-\textbf{value} \\ 
      \midrule
    baseline (number of features: $d=2$) & Age & $0.007_{\pm 0.001}$ & 0.000 \\ 
      \midrule
    \multirow{6}{5cm}{baseline with human-annotated features ($d=2+17$)} 
        & Parenchymal opacification & $-0.194_{\pm 0.081}$ & 0.016 \\ 
      & Pleural effusion & $-0.334_{\pm 0.084}$ & 0.000 \\ 
      & Medical devices & $-0.628_{\pm 0.083}$ & 0.000 \\ 
      & Lung mass & $-0.311_{\pm 0.142}$ & 0.029 \\ 
      & Reticular pattern & $-0.397_{\pm 0.156}$ & 0.011 \\ 
      & Global atelectasis & $-0.430_{\pm 0.216}$ & 0.046 \\ 
        \midrule
     \multirow{12}{5cm}{baseline with algorithm-extracted features ($d=2+76$)} 
        & Shape--Maximum2DDiameterSlice & $-1.547_{\pm 0.512}$ & 0.003 \\ 
      & Firstorder--Entropy & $422.9_{\pm 161.1}$ & 0.009 \\ 
      & GLCM--ID & $-325.9_{\pm 140.0}$ & 0.020 \\ 
      & GLCM--IDN & $1000_{\pm 425.7}$ & 0.019 \\ 
      & GLCM--IMC1 & $296.4_{\pm 85.14}$ & 0.000 \\ 
      & GLCM--InverseVariance & $93.89_{\pm 43.39}$ & 0.030 \\ 
      & GLCM--JointEntropy & $-533.5_{\pm 177.5}$ & 0.003 \\ 
      & GLDM--DependenceEntropy & $156.7_{\pm 57.70}$ & 0.007 \\ 
      & GLDM--LGLE & $-120.9_{\pm 49.07}$ & 0.014 \\ 
      & GLRLM--GLNN & $-497.3_{\pm 207.2}$ & 0.016 \\ 
      & GLRLM--LRE & $6.213_{\pm 2.180}$ & 0.004 \\ 
      & GLSZM--SALGLE & $-694.2_{\pm 283.6}$ & 0.014 \\ 
    \midrule
      \multirow{20}{5cm}{all features ($d=2+17+76$)} 
        & Pleural effusion & $-0.385_{\pm 0.095}$ & 0.000 \\ 
      & Medical devices & $-0.389_{\pm 0.101}$ & 0.000 \\ 
      & Chest wall subcutaneous emphysema & $0.354_{\pm 0.180}$ & 0.049 \\ 
      \cmidrule[0.125pt]{2-4}
      & Shape--Maximum2DDiameterRow & $0.730_{\pm 0.313}$ & 0.020 \\ 
      & Shape--Maximum2DDiameterSlice & $-1.531_{\pm 0.536}$ & 0.004 \\ 
      & Firstorder--Entropy & $407.0_{\pm 165.3}$ & 0.014 \\ 
      & Firstorder--Range & $-4.361_{\pm 1.858}$ & 0.019 \\ 
      & GLCM--ID & $-358.4_{\pm 142.8}$ & 0.012 \\ 
      & GLCM--IDN & $926.6_{\pm 438.3}$ & 0.034 \\ 
      & GLCM--IMC1 & $274.2_{\pm 87.89}$ & 0.002 \\ 
      & GLCM--InverseVariance & $94.82_{\pm 44.86}$ & 0.035 \\ 
      & GLCM--JointEntropy & $-517.9_{\pm 182.2}$ & 0.004 \\ 
      & GLDM--DependenceEntropy & $185.9_{\pm 59.69}$ & 0.002 \\ 
      & GLDM--LGLE & $-136.8_{\pm 50.56}$ & 0.007 \\ 
      & GLRLM--GLNN & $-606.7_{\pm 211.5}$ & 0.004 \\ 
      & GLRLM--LRE & $6.261_{\pm 2.176}$ & 0.004 \\ 
      & GLSZM--SALGLE & $-581.1_{\pm 288.6}$ & 0.044 \\ 
      & NGTDM--Busyness & $1.252_{\pm 0.606}$ & 0.039 \\ 
      & NGTDM--Complexity & $1.406_{\pm 0.625}$ & 0.024 \\ 
       \bottomrule
    \end{tabular}
\end{table*}

\bibliographystyle{bst/elsarticle-num-names} 
\bibliography{references}

\begin{thebibliography}{63}
\expandafter\ifx\csname natexlab\endcsname\relax\def\natexlab#1{#1}\fi
\providecommand{\url}[1]{\texttt{#1}}
\providecommand{\href}[2]{#2}
\providecommand{\path}[1]{#1}
\providecommand{\DOIprefix}{doi:}
\providecommand{\ArXivprefix}{arXiv:}
\providecommand{\URLprefix}{URL: }
\providecommand{\Pubmedprefix}{pmid:}
\providecommand{\doi}[1]{\href{http://dx.doi.org/#1}{\path{#1}}}
\providecommand{\Pubmed}[1]{\href{pmid:#1}{\path{#1}}}
\providecommand{\bibinfo}[2]{#2}
\ifx\xfnm\relax \def\xfnm[#1]{\unskip,\space#1}\fi
\bibitem[{Rajpurkar et~al.(2022)Rajpurkar, Chen, Banerjee, and
  Topol}]{rajpurkar2022ai}
\bibinfo{author}{P.~Rajpurkar}, \bibinfo{author}{E.~Chen},
  \bibinfo{author}{O.~Banerjee}, \bibinfo{author}{E.~J. Topol},
\newblock \bibinfo{title}{{AI} in health and medicine},
\newblock \bibinfo{journal}{Nature Medicine} \bibinfo{volume}{28}
  (\bibinfo{year}{2022}) \bibinfo{pages}{31--38}.
  \DOIprefix\doi{10.1038/s41591-021-01614-0}.
\bibitem[{Zhou et~al.(2023)Zhou, Chia, Wagner, Ayhan, Williamson, Struyven,
  Liu, Xu, Lozano, Woodward-Court et~al.}]{zhou2023foundation}
\bibinfo{author}{Y.~Zhou}, \bibinfo{author}{M.~A. Chia}, \bibinfo{author}{S.~K.
  Wagner}, \bibinfo{author}{M.~S. Ayhan}, \bibinfo{author}{D.~J. Williamson},
  \bibinfo{author}{R.~R. Struyven}, \bibinfo{author}{T.~Liu},
  \bibinfo{author}{M.~Xu}, \bibinfo{author}{M.~G. Lozano},
  \bibinfo{author}{P.~Woodward-Court}, et~al.,
\newblock \bibinfo{title}{A foundation model for generalizable disease
  detection from retinal images},
\newblock \bibinfo{journal}{Nature}  (\bibinfo{year}{2023})
  \bibinfo{pages}{1--8}. \DOIprefix\doi{10.1038/s41586-023-06555-x}.
\bibitem[{Stiglic et~al.(2020)Stiglic, Kocbek, Fijacko, Zitnik, Verbert, and
  Cilar}]{stiglic2020interpretability}
\bibinfo{author}{G.~Stiglic}, \bibinfo{author}{P.~Kocbek},
  \bibinfo{author}{N.~Fijacko}, \bibinfo{author}{M.~Zitnik},
  \bibinfo{author}{K.~Verbert}, \bibinfo{author}{L.~Cilar},
\newblock \bibinfo{title}{Interpretability of machine learning-based prediction
  models in healthcare},
\newblock \bibinfo{journal}{Wiley Interdisciplinary Reviews: Data Mining and
  Knowledge Discovery} \bibinfo{volume}{10} (\bibinfo{year}{2020})
  \bibinfo{pages}{e1379}. \DOIprefix\doi{10.1002/widm.1379}.
\bibitem[{Molnar(2020)}]{molnar2020interpretable}
\bibinfo{author}{C.~Molnar}, \bibinfo{title}{Interpretable Machine Learning --
  A Guide for Making Black Box Models Explainable},
  \bibinfo{publisher}{Independently published}, \bibinfo{year}{2020}.
\bibitem[{Biecek and Burzykowski(2021)}]{biecek2021explanatory}
\bibinfo{author}{P.~Biecek}, \bibinfo{author}{T.~Burzykowski},
  \bibinfo{title}{{Explanatory model analysis: explore, explain, and examine
  predictive models}}, \bibinfo{publisher}{CRC Press}, \bibinfo{year}{2021}.
\bibitem[{Holzinger et~al.(2022)Holzinger, Saranti, Molnar, Biecek, and
  Samek}]{holzinger2022xai}
\bibinfo{author}{A.~Holzinger}, \bibinfo{author}{A.~Saranti},
  \bibinfo{author}{C.~Molnar}, \bibinfo{author}{P.~Biecek},
  \bibinfo{author}{W.~Samek},
\newblock \bibinfo{title}{{Explainable AI Methods -- A Brief Overview}},
\newblock in: \bibinfo{booktitle}{xxAI -- Beyond Explainable AI},
  \bibinfo{year}{2022}, pp. \bibinfo{pages}{13--38}.
  \DOIprefix\doi{10.1007/978-3-031-04083-2_2}.
\bibitem[{Ooge et~al.(2022)Ooge, Stiglic, and Verbert}]{ooge2022explaining}
\bibinfo{author}{J.~Ooge}, \bibinfo{author}{G.~Stiglic},
  \bibinfo{author}{K.~Verbert},
\newblock \bibinfo{title}{Explaining artificial intelligence with visual
  analytics in healthcare},
\newblock \bibinfo{journal}{Wiley Interdisciplinary Reviews: Data Mining and
  Knowledge Discovery} \bibinfo{volume}{12} (\bibinfo{year}{2022})
  \bibinfo{pages}{e1427}. \DOIprefix\doi{10.1002/widm.1427}.
\bibitem[{Combi et~al.(2022)Combi, Amico, Bellazzi, Holzinger, Moore, Zitnik,
  and Holmes}]{combi2022manifesto}
\bibinfo{author}{C.~Combi}, \bibinfo{author}{B.~Amico},
  \bibinfo{author}{R.~Bellazzi}, \bibinfo{author}{A.~Holzinger},
  \bibinfo{author}{J.~H. Moore}, \bibinfo{author}{M.~Zitnik},
  \bibinfo{author}{J.~H. Holmes},
\newblock \bibinfo{title}{A manifesto on explainability for artificial
  intelligence in medicine},
\newblock \bibinfo{journal}{Artificial Intelligence in Medicine}
  \bibinfo{volume}{133} (\bibinfo{year}{2022}) \bibinfo{pages}{102423}.
  \DOIprefix\doi{10.1016/j.artmed.2022.102423}.
\bibitem[{Kovalev et~al.(2020)Kovalev, Utkin, and
  Kasimov}]{kovalev2020survlime}
\bibinfo{author}{M.~S. Kovalev}, \bibinfo{author}{L.~V. Utkin},
  \bibinfo{author}{E.~M. Kasimov},
\newblock \bibinfo{title}{{SurvLIME: A method for explaining machine learning
  survival models}},
\newblock \bibinfo{journal}{{Knowledge-Based Systems}} \bibinfo{volume}{203}
  (\bibinfo{year}{2020}) \bibinfo{pages}{106164}.
  \DOIprefix\doi{10.1016/j.knosys.2020.106164}.
\bibitem[{Wang et~al.(2021)Wang, Samsten, and
  Papapetrou}]{wang2021counterfactual}
\bibinfo{author}{Z.~Wang}, \bibinfo{author}{I.~Samsten},
  \bibinfo{author}{P.~Papapetrou},
\newblock \bibinfo{title}{{Counterfactual Explanations for Survival Prediction
  of Cardiovascular ICU Patients}},
\newblock in: \bibinfo{booktitle}{International Conference on Artificial
  Intelligence in Medicine}, \bibinfo{year}{2021}, pp.
  \bibinfo{pages}{338--348}. \DOIprefix\doi{10.1007/978-3-030-77211-6_38}.
\bibitem[{Rad et~al.(2022)Rad, Tennankore, Vinson, and
  Abidi}]{rad2022extracting}
\bibinfo{author}{J.~Rad}, \bibinfo{author}{K.~K. Tennankore},
  \bibinfo{author}{A.~Vinson}, \bibinfo{author}{S.~S.~R. Abidi},
\newblock \bibinfo{title}{{Extracting Surrogate Decision Trees from Black-Box
  Models to Explain the Temporal Importance of Clinical Features in Predicting
  Kidney Graft Survival}},
\newblock in: \bibinfo{booktitle}{International Conference on Artificial
  Intelligence in Medicine}, \bibinfo{year}{2022}, pp. \bibinfo{pages}{88--98}.
  \DOIprefix\doi{10.1007/978-3-031-09342-5_9}.
\bibitem[{Utkin et~al.(2022)Utkin, Satyukov, and
  Konstantinov}]{utkin2022survnam}
\bibinfo{author}{L.~V. Utkin}, \bibinfo{author}{E.~D. Satyukov},
  \bibinfo{author}{A.~V. Konstantinov},
\newblock \bibinfo{title}{{SurvNAM: The machine learning survival model
  explanation}},
\newblock \bibinfo{journal}{Neural Networks} \bibinfo{volume}{147}
  (\bibinfo{year}{2022}) \bibinfo{pages}{81--102}.
  \DOIprefix\doi{10.1016/j.neunet.2021.12.015}.
\bibitem[{Krzyziński et~al.(2023)Krzyziński, Spytek, Baniecki, and
  Biecek}]{krzyzinski2023survshap}
\bibinfo{author}{M.~Krzyziński}, \bibinfo{author}{M.~Spytek},
  \bibinfo{author}{H.~Baniecki}, \bibinfo{author}{P.~Biecek},
\newblock \bibinfo{title}{{SurvSHAP(t): Time-dependent explanations of machine
  learning survival models}},
\newblock \bibinfo{journal}{Knowledge-Based Systems} \bibinfo{volume}{262}
  (\bibinfo{year}{2023}) \bibinfo{pages}{110234}.
  \DOIprefix\doi{10.1016/j.knosys.2022.110234}.
\bibitem[{Ter-Minassian et~al.(2024)Ter-Minassian, Ghalebikesabi, Diaz-Ordaz,
  and Holmes}]{terminassian2024explainable}
\bibinfo{author}{L.~Ter-Minassian}, \bibinfo{author}{S.~Ghalebikesabi},
  \bibinfo{author}{K.~Diaz-Ordaz}, \bibinfo{author}{C.~Holmes},
\newblock \bibinfo{title}{{Explainable AI for survival analysis: a median-SHAP
  approach}},
\newblock \bibinfo{journal}{arxiv preprint arXiv:2402.00072}
  (\bibinfo{year}{2024}). \DOIprefix\doi{10.48550/arXiv.2402.00072}.
\bibitem[{Langbein et~al.(2024)Langbein, Krzyziński, Spytek, Baniecki, Biecek,
  and Wright}]{langbein2024interpretable}
\bibinfo{author}{S.~H. Langbein}, \bibinfo{author}{M.~Krzyziński},
  \bibinfo{author}{M.~Spytek}, \bibinfo{author}{H.~Baniecki},
  \bibinfo{author}{P.~Biecek}, \bibinfo{author}{M.~N. Wright},
\newblock \bibinfo{title}{{Interpretable Machine Learning for Survival
  Analysis}},
\newblock \bibinfo{journal}{arXiv preprint arXiv:2403.10250}
  (\bibinfo{year}{2024}). \DOIprefix\doi{10.48550/arXiv.2403.10250}.
\bibitem[{Breiman(2001)}]{breiman2001random}
\bibinfo{author}{L.~Breiman},
\newblock \bibinfo{title}{{Random forests}},
\newblock \bibinfo{journal}{Machine learning} \bibinfo{volume}{45}
  (\bibinfo{year}{2001}) \bibinfo{pages}{5--32}.
  \DOIprefix\doi{10.1023/A:1010933404324}.
\bibitem[{Fisher et~al.(2019)Fisher, Rudin, and Dominici}]{fisher2019all}
\bibinfo{author}{A.~Fisher}, \bibinfo{author}{C.~Rudin},
  \bibinfo{author}{F.~Dominici},
\newblock \bibinfo{title}{{All Models are Wrong, but Many are Useful: Learning
  a Variable's Importance by Studying an Entire Class of Prediction Models
  Simultaneously}},
\newblock \bibinfo{journal}{Journal of Machine Learning Research}
  \bibinfo{volume}{20} (\bibinfo{year}{2019}) \bibinfo{pages}{1--81}.
  \DOIprefix\doi{10.48550/arXiv.1801.01489}.
\bibitem[{Goldstein et~al.(2015)Goldstein, Kapelner, Bleich, and
  Pitkin}]{goldstein2015peeking}
\bibinfo{author}{A.~Goldstein}, \bibinfo{author}{A.~Kapelner},
  \bibinfo{author}{J.~Bleich}, \bibinfo{author}{E.~Pitkin},
\newblock \bibinfo{title}{{Peeking inside the black box: Visualizing
  statistical learning with plots of individual conditional expectation}},
\newblock \bibinfo{journal}{Journal of Computational and Graphical Statistics}
  \bibinfo{volume}{24} (\bibinfo{year}{2015}) \bibinfo{pages}{44--65}.
  \DOIprefix\doi{10.1080/10618600.2014.907095}.
\bibitem[{Friedman(2001)}]{friedman2001greedy}
\bibinfo{author}{J.~H. Friedman},
\newblock \bibinfo{title}{{Greedy Function Approximation: A Gradient Boosting
  Machine}},
\newblock \bibinfo{journal}{Annals of Statistics} \bibinfo{volume}{29}
  (\bibinfo{year}{2001}) \bibinfo{pages}{1189--1232}.
  \DOIprefix\doi{10.1214/aos/1013203451}.
\bibitem[{Baniecki et~al.(2023)Baniecki, Sobieski, Bombi{\'n}ski, Szatkowski,
  and Biecek}]{baniecki2023hospital}
\bibinfo{author}{H.~Baniecki}, \bibinfo{author}{B.~Sobieski},
  \bibinfo{author}{P.~Bombi{\'n}ski}, \bibinfo{author}{P.~Szatkowski},
  \bibinfo{author}{P.~Biecek},
\newblock \bibinfo{title}{{Hospital Length of Stay Prediction Based on
  Multi-modal Data Towards Trustworthy Human-AI Collaboration in Radiomics}},
\newblock in: \bibinfo{booktitle}{International Conference on Artificial
  Intelligence in Medicine}, \bibinfo{year}{2023}, pp. \bibinfo{pages}{65--74}.
  \DOIprefix\doi{10.1007/978-3-031-34344-5_9}.
\bibitem[{Pölsterl et~al.(2016)Pölsterl, Conjeti, Navab, and
  Katouzian}]{polsterl2016survival}
\bibinfo{author}{S.~Pölsterl}, \bibinfo{author}{S.~Conjeti},
  \bibinfo{author}{N.~Navab}, \bibinfo{author}{A.~Katouzian},
\newblock \bibinfo{title}{{Survival analysis for high-dimensional,
  heterogeneous medical data: Exploring feature extraction as an alternative to
  feature selection}},
\newblock \bibinfo{journal}{Artificial Intelligence in Medicine}
  \bibinfo{volume}{72} (\bibinfo{year}{2016}) \bibinfo{pages}{1--11}.
  \DOIprefix\doi{10.1016/j.artmed.2016.07.004}.
\bibitem[{Jing et~al.(2019)Jing, Zhang, Wang, Jin, Liu, Qiu, Ke, Sun, He, Hou,
  Tang, Lv, and Li}]{jing2019deep}
\bibinfo{author}{B.~Jing}, \bibinfo{author}{T.~Zhang},
  \bibinfo{author}{Z.~Wang}, \bibinfo{author}{Y.~Jin},
  \bibinfo{author}{K.~Liu}, \bibinfo{author}{W.~Qiu}, \bibinfo{author}{L.~Ke},
  \bibinfo{author}{Y.~Sun}, \bibinfo{author}{C.~He}, \bibinfo{author}{D.~Hou},
  \bibinfo{author}{L.~Tang}, \bibinfo{author}{X.~Lv}, \bibinfo{author}{C.~Li},
\newblock \bibinfo{title}{A deep survival analysis method based on ranking},
\newblock \bibinfo{journal}{Artificial Intelligence in Medicine}
  \bibinfo{volume}{98} (\bibinfo{year}{2019}) \bibinfo{pages}{1--9}.
  \DOIprefix\doi{10.1016/j.artmed.2019.06.001}.
\bibitem[{Hao et~al.(2022)Hao, Li, Feng, Qi, Liu, Arefan, Zhang, and
  Wu}]{hao2022survivalcnn}
\bibinfo{author}{D.~Hao}, \bibinfo{author}{Q.~Li}, \bibinfo{author}{Q.-X.
  Feng}, \bibinfo{author}{L.~Qi}, \bibinfo{author}{X.-S. Liu},
  \bibinfo{author}{D.~Arefan}, \bibinfo{author}{Y.-D. Zhang},
  \bibinfo{author}{S.~Wu},
\newblock \bibinfo{title}{{SurvivalCNN: A deep learning-based method for
  gastric cancer survival prediction using radiological imaging data and
  clinicopathological variables}},
\newblock \bibinfo{journal}{Artificial Intelligence in Medicine}
  \bibinfo{volume}{134} (\bibinfo{year}{2022}) \bibinfo{pages}{102424}.
  \DOIprefix\doi{10.1016/j.artmed.2022.102424}.
\bibitem[{Cho et~al.(2023)Cho, Shu, Bekiranov, Zang, and
  Zhang}]{cho2023interpretable}
\bibinfo{author}{H.~J. Cho}, \bibinfo{author}{M.~Shu},
  \bibinfo{author}{S.~Bekiranov}, \bibinfo{author}{C.~Zang},
  \bibinfo{author}{A.~Zhang},
\newblock \bibinfo{title}{{Interpretable meta-learning of multi-omics data for
  survival analysis and pathway enrichment}},
\newblock \bibinfo{journal}{Bioinformatics} \bibinfo{volume}{39}
  (\bibinfo{year}{2023}) \bibinfo{pages}{btad113}.
  \DOIprefix\doi{10.1093/bioinformatics/btad113}.
\bibitem[{Jiang et~al.(2023)Jiang, Yu, Wang, Ma, and Guan}]{jiang2023decaf}
\bibinfo{author}{J.~Jiang}, \bibinfo{author}{X.~Yu}, \bibinfo{author}{B.~Wang},
  \bibinfo{author}{L.~Ma}, \bibinfo{author}{Y.~Guan},
\newblock \bibinfo{title}{{DECAF: An interpretable deep cascading framework for
  ICU mortality prediction}},
\newblock \bibinfo{journal}{Artificial Intelligence in Medicine}
  \bibinfo{volume}{138} (\bibinfo{year}{2023}) \bibinfo{pages}{102437}.
  \DOIprefix\doi{10.1016/j.artmed.2022.102437}.
\bibitem[{Xu and Guo(2023)}]{xu2023coxnam}
\bibinfo{author}{L.~Xu}, \bibinfo{author}{C.~Guo},
\newblock \bibinfo{title}{{CoxNAM: An interpretable deep survival analysis
  model}},
\newblock \bibinfo{journal}{Expert Systems with Applications}
  \bibinfo{volume}{227} (\bibinfo{year}{2023}) \bibinfo{pages}{120218}.
  \DOIprefix\doi{10.1016/j.eswa.2023.120218}.
\bibitem[{Ribeiro et~al.(2016)Ribeiro, Singh, and Guestrin}]{ribeiro2016should}
\bibinfo{author}{M.~T. Ribeiro}, \bibinfo{author}{S.~Singh},
  \bibinfo{author}{C.~Guestrin},
\newblock \bibinfo{title}{{``Why should I trust you?'' Explaining the
  predictions of any classifier}},
\newblock in: \bibinfo{booktitle}{Proceedings of the 22nd ACM SIGKDD
  international conference on knowledge discovery and data mining},
  \bibinfo{year}{2016}, pp. \bibinfo{pages}{1135--1144}.
  \DOIprefix\doi{10.1145/2939672.2939778}.
\bibitem[{Au et~al.(2022)Au, Herbinger, Stachl, Bischl, and
  Casalicchio}]{au2022grouped}
\bibinfo{author}{Q.~Au}, \bibinfo{author}{J.~Herbinger},
  \bibinfo{author}{C.~Stachl}, \bibinfo{author}{B.~Bischl},
  \bibinfo{author}{G.~Casalicchio},
\newblock \bibinfo{title}{Grouped feature importance and combined features
  effect plot},
\newblock \bibinfo{journal}{Data Mining and Knowledge Discovery}
  \bibinfo{volume}{36} (\bibinfo{year}{2022}) \bibinfo{pages}{1401--1450}.
  \DOIprefix\doi{10.1007/s10618-022-00840-5}.
\bibitem[{Komorowski et~al.(2023)Komorowski, Baniecki, and
  Biecek}]{komorowski2023towards}
\bibinfo{author}{P.~Komorowski}, \bibinfo{author}{H.~Baniecki},
  \bibinfo{author}{P.~Biecek},
\newblock \bibinfo{title}{{Towards Evaluating Explanations of Vision
  Transformers for Medical Imaging}},
\newblock in: \bibinfo{booktitle}{IEEE/CVF Conference on Computer Vision and
  Pattern Recognition Workshops}, \bibinfo{year}{2023}, pp.
  \bibinfo{pages}{3725--3731}. \DOIprefix\doi{10.1109/CVPRW59228.2023.00383}.
\bibitem[{Donizy et~al.(2023)Donizy, Spytek, Krzyzi{\'n}ski, Kotowski,
  Markiewicz, Romanowska-Dixon, Biecek, and Hoang}]{donizy2023ki67}
\bibinfo{author}{P.~Donizy}, \bibinfo{author}{M.~Spytek},
  \bibinfo{author}{M.~Krzyzi{\'n}ski}, \bibinfo{author}{K.~Kotowski},
  \bibinfo{author}{A.~Markiewicz}, \bibinfo{author}{B.~Romanowska-Dixon},
  \bibinfo{author}{P.~Biecek}, \bibinfo{author}{M.~P. Hoang},
\newblock \bibinfo{title}{{Ki67 is a better marker than PRAME in risk
  stratification of BAP1-positive and BAP1-loss uveal melanomas}},
\newblock \bibinfo{journal}{British Journal of Ophthalmology}
  (\bibinfo{year}{2023}). \DOIprefix\doi{10.1136/bjo-2023-323816}.
\bibitem[{Baniecki et~al.(2023)Baniecki, Parzych, and
  Biecek}]{baniecki2023grammar}
\bibinfo{author}{H.~Baniecki}, \bibinfo{author}{D.~Parzych},
  \bibinfo{author}{P.~Biecek},
\newblock \bibinfo{title}{The grammar of interactive explanatory model
  analysis},
\newblock \bibinfo{journal}{Data Mining and Knowledge Discovery}
  (\bibinfo{year}{2023}) \bibinfo{pages}{1--37}.
  \DOIprefix\doi{10.1007/s10618-023-00924-w}.
\bibitem[{Huang et~al.(2013)Huang, Juarez, Duan, and Li}]{huang2013length}
\bibinfo{author}{Z.~Huang}, \bibinfo{author}{J.~M. Juarez},
  \bibinfo{author}{H.~Duan}, \bibinfo{author}{H.~Li},
\newblock \bibinfo{title}{{Length of stay prediction for clinical treatment
  process using temporal similarity}},
\newblock \bibinfo{journal}{Expert Systems with Applications}
  \bibinfo{volume}{40} (\bibinfo{year}{2013}) \bibinfo{pages}{6330--6339}.
  \DOIprefix\doi{10.1016/j.eswa.2013.05.066}.
\bibitem[{Chaou et~al.(2017)Chaou, Chen, Chang, Tang, Pan, Yen, and
  Chiu}]{chaou2017predicting}
\bibinfo{author}{C.-H. Chaou}, \bibinfo{author}{H.-H. Chen},
  \bibinfo{author}{S.-H. Chang}, \bibinfo{author}{P.~Tang},
  \bibinfo{author}{S.-L. Pan}, \bibinfo{author}{A.~M.-F. Yen},
  \bibinfo{author}{T.-F. Chiu},
\newblock \bibinfo{title}{Predicting length of stay among patients discharged
  from the emergency department—using an accelerated failure time model},
\newblock \bibinfo{journal}{PloS one} \bibinfo{volume}{12}
  (\bibinfo{year}{2017}) \bibinfo{pages}{e0165756}.
  \DOIprefix\doi{10.1371/journal.pone.0165756}.
\bibitem[{Rudin(2019)}]{rudin2019stop}
\bibinfo{author}{C.~Rudin},
\newblock \bibinfo{title}{{Stop Explaining Black Box Machine Learning Models
  for High Stakes Decisions and Use Interpretable Models Instead}},
\newblock \bibinfo{journal}{Nature Machine Intelligence} \bibinfo{volume}{1}
  (\bibinfo{year}{2019}) \bibinfo{pages}{206--215}.
  \DOIprefix\doi{10.1038/s42256-019-0048-x}.
\bibitem[{Muhlestein et~al.(2019)Muhlestein, Akagi, Davies, and
  Chambless}]{muhlestein2019predicting}
\bibinfo{author}{W.~E. Muhlestein}, \bibinfo{author}{D.~S. Akagi},
  \bibinfo{author}{J.~M. Davies}, \bibinfo{author}{L.~B. Chambless},
\newblock \bibinfo{title}{{Predicting inpatient length of stay after brain
  tumor surgery: developing machine learning ensembles to improve predictive
  performance}},
\newblock \bibinfo{journal}{Neurosurgery} \bibinfo{volume}{85}
  (\bibinfo{year}{2019}) \bibinfo{pages}{384--393}.
  \DOIprefix\doi{10.1186/s12911-022-01855-0}.
\bibitem[{Zhang et~al.(2020)Zhang, Yin, Zeng, Yuan, and
  Zhang}]{zhang2020combining}
\bibinfo{author}{D.~Zhang}, \bibinfo{author}{C.~Yin},
  \bibinfo{author}{J.~Zeng}, \bibinfo{author}{X.~Yuan},
  \bibinfo{author}{P.~Zhang},
\newblock \bibinfo{title}{{Combining structured and unstructured data for
  predictive models: a deep learning approach}},
\newblock \bibinfo{journal}{BMC Medical Informatics and Decision Making}
  \bibinfo{volume}{20} (\bibinfo{year}{2020}) \bibinfo{pages}{1--11}.
  \DOIprefix\doi{10.1186/s12911-020-01297-6}.
\bibitem[{Wen et~al.(2022)Wen, Rahman, Zhuang, Pokojovy, Xu, McCaffrey, Vo,
  Walser, Moen, and Tseng}]{wen2022time}
\bibinfo{author}{Y.~Wen}, \bibinfo{author}{M.~F. Rahman},
  \bibinfo{author}{Y.~Zhuang}, \bibinfo{author}{M.~Pokojovy},
  \bibinfo{author}{H.~Xu}, \bibinfo{author}{P.~McCaffrey},
  \bibinfo{author}{A.~Vo}, \bibinfo{author}{E.~Walser},
  \bibinfo{author}{S.~Moen}, \bibinfo{author}{T.-L. Tseng},
\newblock \bibinfo{title}{{Time-to-event modeling for hospital length of stay
  prediction for COVID-19 patients}},
\newblock \bibinfo{journal}{Machine Learning with Applications}
  \bibinfo{volume}{9} (\bibinfo{year}{2022}) \bibinfo{pages}{100365}.
  \DOIprefix\doi{10.1016/j.mlwa.2022.100365}.
\bibitem[{Stone et~al.(2022)Stone, Zwiggelaar, Jones, and
  Mac~Parthal{\'a}in}]{stone2022systematic}
\bibinfo{author}{K.~Stone}, \bibinfo{author}{R.~Zwiggelaar},
  \bibinfo{author}{P.~Jones}, \bibinfo{author}{N.~Mac~Parthal{\'a}in},
\newblock \bibinfo{title}{{A systematic review of the prediction of hospital
  length of stay: Towards a unified framework}},
\newblock \bibinfo{journal}{PLOS Digital Health} \bibinfo{volume}{1}
  (\bibinfo{year}{2022}) \bibinfo{pages}{e0000017}.
  \DOIprefix\doi{10.1371/journal.pdig.0000017}.
\bibitem[{Van~Griethuysen et~al.(2017)Van~Griethuysen, Fedorov, Parmar
  et~al.}]{van2017computational}
\bibinfo{author}{J.~J. Van~Griethuysen}, \bibinfo{author}{A.~Fedorov},
  \bibinfo{author}{C.~Parmar}, et~al.,
\newblock \bibinfo{title}{{Computational radiomics system to decode the
  radiographic phenotype}},
\newblock \bibinfo{journal}{Cancer research} \bibinfo{volume}{77}
  (\bibinfo{year}{2017}) \bibinfo{pages}{e104--e107}.
  \DOIprefix\doi{10.1158/0008-5472.CAN-17-0339}.
\bibitem[{Spytek et~al.(2023)Spytek, Krzyziński, Langbein, Baniecki, Wright,
  and Biecek}]{spytek2023survex}
\bibinfo{author}{M.~Spytek}, \bibinfo{author}{M.~Krzyziński},
  \bibinfo{author}{S.~H. Langbein}, \bibinfo{author}{H.~Baniecki},
  \bibinfo{author}{M.~N. Wright}, \bibinfo{author}{P.~Biecek},
\newblock \bibinfo{title}{{survex: an R package for explaining machine learning
  survival models}},
\newblock \bibinfo{journal}{Bioinformatics} \bibinfo{volume}{39}
  (\bibinfo{year}{2023}) \bibinfo{pages}{btad723}.
  \DOIprefix\doi{10.1093/bioinformatics/btad723}.
\bibitem[{Apley and Zhu(2020)}]{apley2020visualizing}
\bibinfo{author}{D.~W. Apley}, \bibinfo{author}{J.~Zhu},
\newblock \bibinfo{title}{{Visualizing the effects of predictor variables in
  black box supervised learning models}},
\newblock \bibinfo{journal}{Journal of the Royal Statistical Society: Series B
  (Statistical Methodology)} \bibinfo{volume}{82} (\bibinfo{year}{2020})
  \bibinfo{pages}{1059--1086}. \DOIprefix\doi{10.1111/rssb.12377}.
\bibitem[{Gkolemis et~al.(2023)Gkolemis, Dalamagas, and
  Diou}]{gkolemis2023dale}
\bibinfo{author}{V.~Gkolemis}, \bibinfo{author}{T.~Dalamagas},
  \bibinfo{author}{C.~Diou},
\newblock \bibinfo{title}{{DALE: Differential Accumulated Local Effects for
  efficient and accurate global explanations}},
\newblock in: \bibinfo{booktitle}{Asian Conference on Machine Learning}, volume
  \bibinfo{volume}{189}, \bibinfo{year}{2023}, pp. \bibinfo{pages}{375--390}.
  \DOIprefix\doi{10.48550/arXiv.2210.04542}.
\bibitem[{Covert et~al.(2020)Covert, Lundberg, and
  Lee}]{covert2020understanding}
\bibinfo{author}{I.~Covert}, \bibinfo{author}{S.~Lundberg},
  \bibinfo{author}{S.-I. Lee},
\newblock \bibinfo{title}{{Understanding Global Feature Contributions With
  Additive Importance Measures}},
\newblock in: \bibinfo{booktitle}{Advances in Neural Information Processing
  Systems}, \bibinfo{year}{2020}, pp. \bibinfo{pages}{17212--17223}.
  \DOIprefix\doi{10.48550/arXiv.2004.00668}.
\bibitem[{Gu et~al.(2019)Gu, Cheng, Fu, Zhou, Hao, Zhao, Zhang, Gao, and
  Liu}]{gu2019net}
\bibinfo{author}{Z.~Gu}, \bibinfo{author}{J.~Cheng}, \bibinfo{author}{H.~Fu},
  \bibinfo{author}{K.~Zhou}, \bibinfo{author}{H.~Hao},
  \bibinfo{author}{Y.~Zhao}, \bibinfo{author}{T.~Zhang},
  \bibinfo{author}{S.~Gao}, \bibinfo{author}{J.~Liu},
\newblock \bibinfo{title}{{CE-Net: Context encoder network for 2D medical image
  segmentation}},
\newblock \bibinfo{journal}{IEEE Transactions on Medical Imaging}
  \bibinfo{volume}{38} (\bibinfo{year}{2019}) \bibinfo{pages}{2281--2292}.
  \DOIprefix\doi{10.1109/TMI.2019.2903562}.
\bibitem[{Hansell et~al.(2008)Hansell, Bankier, MacMahon, McLoud, Muller, Remy
  et~al.}]{hansell2008fleischner}
\bibinfo{author}{D.~M. Hansell}, \bibinfo{author}{A.~A. Bankier},
  \bibinfo{author}{H.~MacMahon}, \bibinfo{author}{T.~C. McLoud},
  \bibinfo{author}{N.~L. Muller}, \bibinfo{author}{J.~Remy}, et~al.,
\newblock \bibinfo{title}{{Fleischner Society: glossary of terms for thoracic
  imaging}},
\newblock \bibinfo{journal}{Radiology} \bibinfo{volume}{246}
  (\bibinfo{year}{2008}) \bibinfo{pages}{697}.
  \DOIprefix\doi{10.1148/radiol.2462070712}.
\bibitem[{{Radiological Society of North America}(2023)}]{radlex}
\bibinfo{author}{{Radiological Society of North America}},
  \bibinfo{title}{{Radiology Lexicon}}, \bibinfo{year}{2023}.
  \bibinfo{note}{\url{https://radlex.org}}.
\bibitem[{Irvin et~al.(2019)Irvin, Rajpurkar, Ko, Yu, Ciurea-Ilcus, Chute,
  Marklund, Haghgoo, Ball, Shpanskaya, Seekins, Mong, Halabi, Sandberg, Jones,
  Larson, Langlotz, Patel, Lungren, and Ng}]{irvin2019chexpert}
\bibinfo{author}{J.~Irvin}, \bibinfo{author}{P.~Rajpurkar},
  \bibinfo{author}{M.~Ko}, \bibinfo{author}{Y.~Yu},
  \bibinfo{author}{S.~Ciurea-Ilcus}, \bibinfo{author}{C.~Chute},
  \bibinfo{author}{H.~Marklund}, \bibinfo{author}{B.~Haghgoo},
  \bibinfo{author}{R.~Ball}, \bibinfo{author}{K.~Shpanskaya},
  \bibinfo{author}{J.~Seekins}, \bibinfo{author}{D.~A. Mong},
  \bibinfo{author}{S.~S. Halabi}, \bibinfo{author}{J.~K. Sandberg},
  \bibinfo{author}{R.~Jones}, \bibinfo{author}{D.~B. Larson},
  \bibinfo{author}{C.~P. Langlotz}, \bibinfo{author}{B.~N. Patel},
  \bibinfo{author}{M.~P. Lungren}, \bibinfo{author}{A.~Y. Ng},
\newblock \bibinfo{title}{{CheXpert: A Large Chest Radiograph Dataset with
  Uncertainty Labels and Expert Comparison}},
\newblock in: \bibinfo{booktitle}{Proceedings of the AAAI Conference on
  Artificial Intelligence}, volume~\bibinfo{volume}{33}, \bibinfo{year}{2019},
  p. \bibinfo{pages}{590–597}. \DOIprefix\doi{10.1609/aaai.v33i01.3301590}.
\bibitem[{Johnson et~al.(2023)Johnson, Bulgarelli, Shen, Gayles, Shammout,
  Horng, Pollard, Hao, Moody, Gow et~al.}]{johnson2023mimic}
\bibinfo{author}{A.~E. Johnson}, \bibinfo{author}{L.~Bulgarelli},
  \bibinfo{author}{L.~Shen}, \bibinfo{author}{A.~Gayles},
  \bibinfo{author}{A.~Shammout}, \bibinfo{author}{S.~Horng},
  \bibinfo{author}{T.~J. Pollard}, \bibinfo{author}{S.~Hao},
  \bibinfo{author}{B.~Moody}, \bibinfo{author}{B.~Gow}, et~al.,
\newblock \bibinfo{title}{{MIMIC-IV, a freely accessible electronic health
  record dataset}},
\newblock \bibinfo{journal}{Scientific data} \bibinfo{volume}{10}
  (\bibinfo{year}{2023}) \bibinfo{pages}{1}.
  \DOIprefix\doi{10.1038/s41597-022-01899-x}.
\bibitem[{Nguyen et~al.(2022)Nguyen, Lam, Le, Pham, Tran, Nguyen, Le, Pham,
  Tong, Dinh et~al.}]{nguyen2022vindr}
\bibinfo{author}{H.~Q. Nguyen}, \bibinfo{author}{K.~Lam},
  \bibinfo{author}{L.~T. Le}, \bibinfo{author}{H.~H. Pham},
  \bibinfo{author}{D.~Q. Tran}, \bibinfo{author}{D.~B. Nguyen},
  \bibinfo{author}{D.~D. Le}, \bibinfo{author}{C.~M. Pham},
  \bibinfo{author}{H.~T. Tong}, \bibinfo{author}{D.~H. Dinh}, et~al.,
\newblock \bibinfo{title}{{VinDr-CXR: An open dataset of chest X-rays with
  radiologist’s annotations}},
\newblock \bibinfo{journal}{Scientific Data} \bibinfo{volume}{9}
  (\bibinfo{year}{2022}) \bibinfo{pages}{429}.
  \DOIprefix\doi{10.1038/s41597-022-01498-w}.
\bibitem[{Sonabend et~al.(2021)Sonabend, Király, Bender, Bischl, and
  Lang}]{sonabend2021mlr3proba}
\bibinfo{author}{R.~Sonabend}, \bibinfo{author}{F.~J. Király},
  \bibinfo{author}{A.~Bender}, \bibinfo{author}{B.~Bischl},
  \bibinfo{author}{M.~Lang},
\newblock \bibinfo{title}{{mlr3proba: An R Package for Machine Learning in
  Survival Analysis}},
\newblock \bibinfo{journal}{Bioinformatics}  (\bibinfo{year}{2021})
  \bibinfo{pages}{2789--2791}. \DOIprefix\doi{10.1093/bioinformatics/btab039}.
\bibitem[{Herrmann et~al.(2021{\natexlab{a}})Herrmann, Probst, Hornung,
  Jurinovic, and Boulesteix}]{herrman2021benchmark}
\bibinfo{author}{M.~Herrmann}, \bibinfo{author}{P.~Probst},
  \bibinfo{author}{R.~Hornung}, \bibinfo{author}{V.~Jurinovic},
  \bibinfo{author}{A.-L. Boulesteix},
\newblock \bibinfo{title}{{Large-scale benchmark study of survival prediction
  methods using multi-omics data}},
\newblock \bibinfo{journal}{Briefings in Bioinformatics} \bibinfo{volume}{22}
  (\bibinfo{year}{2021}{\natexlab{a}}). \DOIprefix\doi{10.1093/bib/bbaa167}.
\bibitem[{Herrmann et~al.(2021{\natexlab{b}})Herrmann, Probst, Hornung,
  Jurinovic, and Boulesteix}]{herrmann2021largescale}
\bibinfo{author}{M.~Herrmann}, \bibinfo{author}{P.~Probst},
  \bibinfo{author}{R.~Hornung}, \bibinfo{author}{V.~Jurinovic},
  \bibinfo{author}{A.-L. Boulesteix},
\newblock \bibinfo{title}{{Large-scale benchmark study of survival prediction
  methods using multi-omics data}},
\newblock \bibinfo{journal}{Briefings in Bioinformatics} \bibinfo{volume}{22}
  (\bibinfo{year}{2021}{\natexlab{b}}) \bibinfo{pages}{bbaa167}.
  \DOIprefix\doi{10.1093/bib/bbaa167}.
\bibitem[{Bommert et~al.(2022)Bommert, Welchowski, Schmid, and
  Rahnenführer}]{bommert2022benchmark}
\bibinfo{author}{A.~Bommert}, \bibinfo{author}{T.~Welchowski},
  \bibinfo{author}{M.~Schmid}, \bibinfo{author}{J.~Rahnenführer},
\newblock \bibinfo{title}{{Benchmark of filter methods for feature selection in
  high-dimensional gene expression survival data}},
\newblock \bibinfo{journal}{Briefings in Bioinformatics} \bibinfo{volume}{23}
  (\bibinfo{year}{2022}) \bibinfo{pages}{bbab354}.
  \DOIprefix\doi{10.1093/bib/bbab354}.
\bibitem[{Gichoya et~al.(2022)Gichoya, Banerjee, Bhimireddy, Burns, Celi, Chen,
  Correa, Dullerud, Ghassemi, Huang, Kuo, Lungren, Palmer, Price, Purkayastha,
  Pyrros, Oakden-Rayner, Okechukwu, Seyyed-Kalantari, Trivedi, Wang, Zaiman,
  and Zhang}]{gichoya2022ai}
\bibinfo{author}{J.~W. Gichoya}, \bibinfo{author}{I.~Banerjee},
  \bibinfo{author}{A.~R. Bhimireddy}, \bibinfo{author}{J.~L. Burns},
  \bibinfo{author}{L.~A. Celi}, \bibinfo{author}{L.-C. Chen},
  \bibinfo{author}{R.~Correa}, \bibinfo{author}{N.~Dullerud},
  \bibinfo{author}{M.~Ghassemi}, \bibinfo{author}{S.-C. Huang},
  \bibinfo{author}{P.-C. Kuo}, \bibinfo{author}{M.~P. Lungren},
  \bibinfo{author}{L.~J. Palmer}, \bibinfo{author}{B.~J. Price},
  \bibinfo{author}{S.~Purkayastha}, \bibinfo{author}{A.~T. Pyrros},
  \bibinfo{author}{L.~Oakden-Rayner}, \bibinfo{author}{C.~Okechukwu},
  \bibinfo{author}{L.~Seyyed-Kalantari}, \bibinfo{author}{H.~Trivedi},
  \bibinfo{author}{R.~Wang}, \bibinfo{author}{Z.~Zaiman},
  \bibinfo{author}{H.~Zhang},
\newblock \bibinfo{title}{{AI} recognition of patient race in medical imaging:
  a modelling study},
\newblock \bibinfo{journal}{The Lancet Digital Health} \bibinfo{volume}{4}
  (\bibinfo{year}{2022}) \bibinfo{pages}{e406--e414}.
  \DOIprefix\doi{10.1016/S2589-7500(22)00063-2}.
\bibitem[{Molnar et~al.(2022)Molnar, K{\"o}nig, Herbinger, Freiesleben, Dandl,
  Scholbeck, Casalicchio, Grosse-Wentrup, and Bischl}]{molnar2020general}
\bibinfo{author}{C.~Molnar}, \bibinfo{author}{G.~K{\"o}nig},
  \bibinfo{author}{J.~Herbinger}, \bibinfo{author}{T.~Freiesleben},
  \bibinfo{author}{S.~Dandl}, \bibinfo{author}{C.~A. Scholbeck},
  \bibinfo{author}{G.~Casalicchio}, \bibinfo{author}{M.~Grosse-Wentrup},
  \bibinfo{author}{B.~Bischl},
\newblock \bibinfo{title}{General pitfalls of model-agnostic interpretation
  methods for machine learning models},
\newblock in: \bibinfo{booktitle}{xxAI -- Beyond Explainable AI},
  \bibinfo{year}{2022}, pp. \bibinfo{pages}{39--68}.
  \DOIprefix\doi{10.1007/978-3-031-04083-2_4}.
\bibitem[{Molnar et~al.(2023)Molnar, K{\"o}nig, Bischl, and
  Casalicchio}]{molnar2023model}
\bibinfo{author}{C.~Molnar}, \bibinfo{author}{G.~K{\"o}nig},
  \bibinfo{author}{B.~Bischl}, \bibinfo{author}{G.~Casalicchio},
\newblock \bibinfo{title}{Model-agnostic feature importance and effects with
  dependent features: a conditional subgroup approach},
\newblock \bibinfo{journal}{Data Mining and Knowledge Discovery}
  (\bibinfo{year}{2023}) \bibinfo{pages}{1--39}.
  \DOIprefix\doi{10.1007/978-3-031-04083-2_4}.
\bibitem[{Aas et~al.(2021)Aas, Jullum, and Løland}]{aas2021explaining}
\bibinfo{author}{K.~Aas}, \bibinfo{author}{M.~Jullum},
  \bibinfo{author}{A.~Løland},
\newblock \bibinfo{title}{Explaining individual predictions when features are
  dependent: More accurate approximations to shapley values},
\newblock \bibinfo{journal}{Artificial Intelligence} \bibinfo{volume}{298}
  (\bibinfo{year}{2021}) \bibinfo{pages}{103502}.
  \DOIprefix\doi{10.1016/j.artint.2021.103502}.
\bibitem[{Turb{\'e} et~al.(2019)Turb{\'e}, Bjelogrlic, Lovis, and
  Mengaldo}]{turbe2023evaluation}
\bibinfo{author}{H.~Turb{\'e}}, \bibinfo{author}{M.~Bjelogrlic},
  \bibinfo{author}{C.~Lovis}, \bibinfo{author}{G.~Mengaldo},
\newblock \bibinfo{title}{{Evaluation of post-hoc interpretability methods in
  time-series classificationd}},
\newblock \bibinfo{journal}{Nature Machine Intelligence} \bibinfo{volume}{1}
  (\bibinfo{year}{2019}) \bibinfo{pages}{206--215}.
  \DOIprefix\doi{10.1038/s42256-023-00620-w}.
\bibitem[{Baniecki and Biecek(2024)}]{baniecki2024advxai}
\bibinfo{author}{H.~Baniecki}, \bibinfo{author}{P.~Biecek},
\newblock \bibinfo{title}{Adversarial attacks and defenses in explainable
  artificial intelligence: A survey},
\newblock \bibinfo{journal}{Information Fusion} \bibinfo{volume}{107}
  (\bibinfo{year}{2024}) \bibinfo{pages}{102303}.
  \DOIprefix\doi{10.1016/j.inffus.2024.102303}.
\bibitem[{Noppel and Wressnegger(2024)}]{noppel2024explainable}
\bibinfo{author}{M.~Noppel}, \bibinfo{author}{C.~Wressnegger},
\newblock \bibinfo{title}{{SoK: Explainable Machine Learning in Adversarial
  Environments}},
\newblock in: \bibinfo{booktitle}{IEEE Symposium on Security and Privacy},
  \bibinfo{year}{2024}, p. \bibinfo{pages}{in press}.
  \DOIprefix\doi{10.1109/SP54263.2024.00021}.
\bibitem[{Poursabzi-Sangdeh et~al.(2021)Poursabzi-Sangdeh, Goldstein, Hofman,
  Wortman~Vaughan, and Wallach}]{poursabzi2021manipulating}
\bibinfo{author}{F.~Poursabzi-Sangdeh}, \bibinfo{author}{D.~G. Goldstein},
  \bibinfo{author}{J.~M. Hofman}, \bibinfo{author}{J.~W. Wortman~Vaughan},
  \bibinfo{author}{H.~Wallach},
\newblock \bibinfo{title}{{Manipulating and Measuring Model Interpretability}},
\newblock in: \bibinfo{booktitle}{CHI Conference on Human Factors in Computing
  Systems}, \bibinfo{year}{2021}, pp. \bibinfo{pages}{1--52}.
  \DOIprefix\doi{10.1145/3411764.3445315}.
\bibitem[{Chen et~al.(2023)Chen, Covert, Lundberg, and
  Lee}]{chen2023algorithms}
\bibinfo{author}{H.~Chen}, \bibinfo{author}{I.~C. Covert},
  \bibinfo{author}{S.~M. Lundberg}, \bibinfo{author}{S.-I. Lee},
\newblock \bibinfo{title}{Algorithms to estimate {Shapley} value feature
  attributions},
\newblock \bibinfo{journal}{Nature Machine Intelligence} \bibinfo{volume}{5}
  (\bibinfo{year}{2023}) \bibinfo{pages}{590--601}.
  \DOIprefix\doi{10.1038/s42256-023-00657-x}.
\bibitem[{Bischl et~al.(2021)Bischl, Casalicchio, Feurer, Gijsbers, Hutter,
  Lang, Mantovani, van Rijn, and Vanschoren}]{bischl2021openml}
\bibinfo{author}{B.~Bischl}, \bibinfo{author}{G.~Casalicchio},
  \bibinfo{author}{M.~Feurer}, \bibinfo{author}{P.~Gijsbers},
  \bibinfo{author}{F.~Hutter}, \bibinfo{author}{M.~Lang},
  \bibinfo{author}{R.~G. Mantovani}, \bibinfo{author}{J.~N. van Rijn},
  \bibinfo{author}{J.~Vanschoren},
\newblock \bibinfo{title}{{OpenML} benchmarking suites},
\newblock in: \bibinfo{booktitle}{Advances in Neural Information Processing
  Systems}, \bibinfo{year}{2021}. \DOIprefix\doi{10.48550/arXiv.1708.03731}.

\end{thebibliography}

\end{document}